\pdfoutput=1
\documentclass[11pt]{article}

\usepackage[preprint]{acl}

\usepackage{times}
\usepackage{latexsym}

\usepackage[T1]{fontenc}
\usepackage[utf8]{inputenc}

\usepackage{microtype}

\usepackage{inconsolata}

\usepackage[table]{xcolor}
\usepackage{algorithm}
\usepackage{algpseudocode}
\usepackage{graphicx}
\usepackage{amsmath}
\usepackage{amssymb}
\usepackage{booktabs}
\usepackage{multirow}
\usepackage{arydshln}
\usepackage{minted}
\usepackage{fvextra} % Provides extra features for verbatim environments like line breaking
\DefineVerbatimEnvironment{Verbatim}{Verbatim}{breaklines=true}
\usepackage{makecell}
\usepackage{pifont}
\usepackage{array}
\usepackage{varwidth}
\usepackage{enumitem}
\usepackage{multirow}
\usepackage{caption}
\usepackage{makecell}
\usepackage{svg}
\usepackage{float}
\usepackage{pdfpages}
\usepackage{colortbl}
\usepackage{tikz,pgfplots}
\usepackage{listings}
\usepackage{tablefootnote}
\usepackage{pythonhighlight}
\usepackage{tabularx}
\usepackage{placeins}
\usepackage{booktabs}
\usepackage{tabularx}
\usepackage{array}
\usepackage{float}

\newcommand{\bad}[1]{\textcolor{red}{#1}}
\newcommand{\good}[1]{\textcolor{green!60!black}{#1}}
\newcommand{\badmath}[1]{{\color{red}#1}}
\newcommand{\goodmath}[1]{{\color{green!60!black}#1}}

\usepackage{ragged2e}
\newcolumntype{Y}{>{\raggedright\arraybackslash}X}

\usepackage{adjustbox}
\usepackage[most]{tcolorbox}
\tcbset{
  colback=gray!3,
  colframe=gray!40,
  boxrule=0.4pt,
  arc=2pt,
  left=4pt,right=4pt,top=3pt,bottom=3pt,
  breakable,
  enhanced,
  before skip=0pt,
  after skip=0pt
}

\usepackage{enumitem}
\setlist[itemize]{leftmargin=*, nosep}

\newcolumntype{Y}{>{\centering\arraybackslash}X}

\title{FLEx: Language Modeling with \textbf{F}ew-shot \textbf{L}anguage \textbf{Ex}planations}
\author{
    Adar Avsian$^{1*}$,
    Christopher Richardson$^{1*}$,
    Anirudh Sundar $^{2}$\thanks{Work done while at Georgia Tech},
    Larry Heck $^{1}$\\
    $^{1}$ Georgia Institute of Technology \\
  $^{2}$ Microsoft\\
  $^{*}$ Equal Contribution \\
  \texttt{\{adar, crichardson8, larryheck\}@gatech.edu} \quad
  \texttt{anisundar@microsoft.com} \quad
}

\usepackage{amsmath}

\definecolor{lightlightgray}{gray}{0.9} 

\begin{document}
\maketitle
\begin{abstract}
Language models have become effective at a wide range of tasks, from math problem solving to open-domain question answering. However, they still make mistakes, and these mistakes are often repeated across related queries. Natural language explanations can help correct these errors, but collecting them at scale may be infeasible, particularly in domains where expert annotators are required. To address this issue, we introduce FLEx (\textbf{F}ew-shot \textbf{L}anguage \textbf{Ex}planations), a method for improving model behavior using a small number of explanatory examples. FLEx selects representative model errors using embedding-based clustering, verifies that the associated explanations correct those errors, and summarizes them into a prompt prefix that is prepended at inference-time. This summary guides the model to avoid similar errors on new inputs, without modifying model weights. We evaluate FLEx on CounterBench, GSM8K, and ReasonIF. We find that FLEx consistently outperforms chain-of-thought (CoT) prompting across all three datasets and reduces up to 83\% of CoT's remaining errors.

\end{abstract}

\section{Introduction}

Large Language Models (LLMs) have advanced rapidly in recent years. While the Transformer architecture \cite{vaswani2017attention} provided the architectural foundation of modern language models, large-scale pretraining like that introduced with GPT \cite{radford2018improving} and BERT \cite{devlin2018bert} enabled language models to be multitask learners, and the billion-parameter scaling of GPT-3 \cite{brown2020language} brought about the era of few-shot learning models with emergent capabilities. InstructGPT \cite{ouyang2022training} introduced instruction tuning, which gave rise to the current generation of multitask, prompt-following language models.

Despite these developments, LLMs remain prone to systematic errors. Because these models are static at inference-time, they often repeat similar mistakes across related queries. A model that misinterprets a counterfactual statement or applies an incorrect strategy on one query will often repeat the same mistake on closely related ones. While prompting strategies like chain-of-thought reasoning \cite{wei2022chain} can improve reasoning and accuracy, they do not reliably prevent recurring errors or ensure correctness \cite{lanham2023measuring, turpin2023language}.

A growing body of work attempts to address these issues by incorporating explanations: natural language descriptions of why an answer is correct or why an error occurred \cite{zelikman2022star, chen2024learning, shinn2023reFLExion, madaan2023self, huang2022large, wu2024discret, lampinen2022can, richardson2023syndicom}. While effective, these approaches typically require large numbers of annotated explanations, multi-step interaction, or model parameter updates, making them expensive or difficult to deploy in settings with limited expert supervision.

\begin{figure*}[t]
    \centering
    \includegraphics[width=\linewidth]{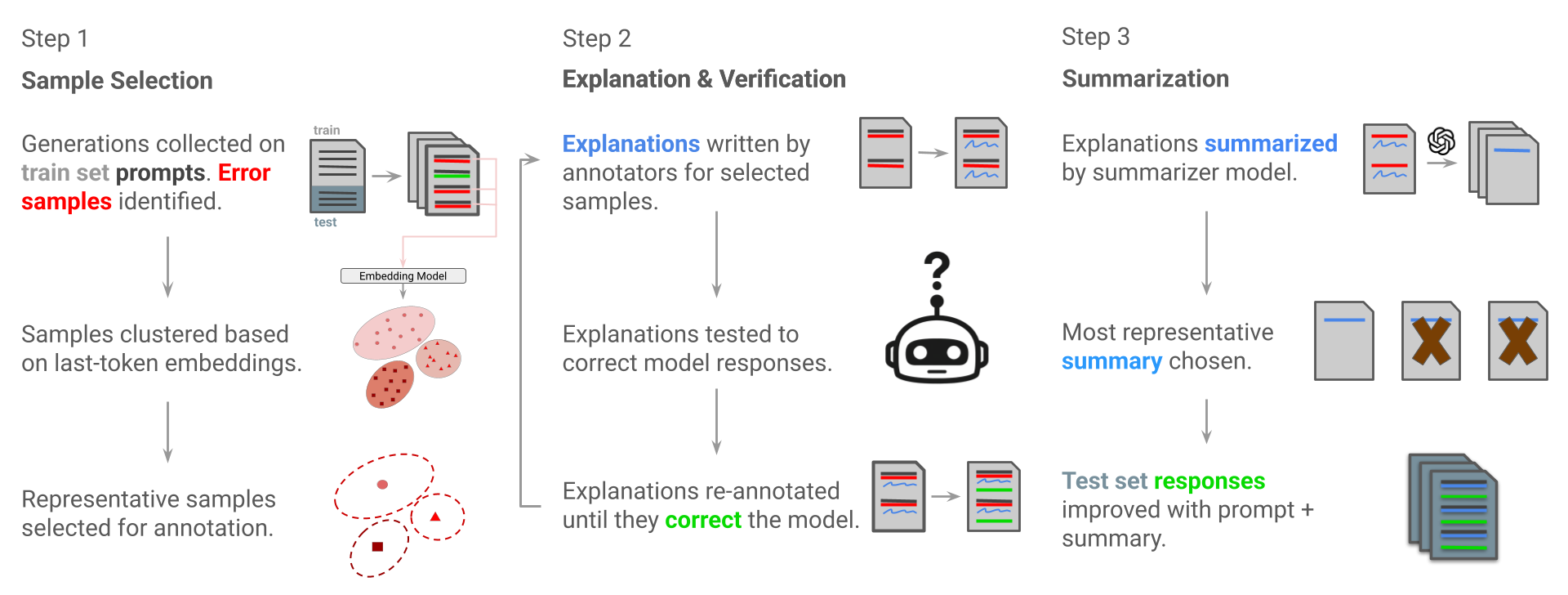}
    \caption{Visualization of the FLEx method for learning from few-shot explanatory explanations. Black bars indicate the prompts, blue bars the summaries, red bars incorrect responses, and green bars correct responses.}
    \label{fig:flex}
\end{figure*}

In this work, we explore the use of human-written natural language explanations to help LLMs generalize away from recurring errors without the need for annotation at scale. To this end, we propose \textbf{FLEx} (\textit{\textbf{F}ew-shot \textbf{L}anguage \textbf{Ex}planations}). As illustrated in Figure \ref{fig:flex}, FLEx operates in three steps. First, using model generations on the training split, we identify a relatively small number of representative error cases via embedding-based clustering. Second, annotators provide natural language explanations for these errors, which are iteratively verified on the training split by re-evaluating the model until the explanation corrects the model's response. Third, we distill the verified explanations into a concise prompt prefix using a summary selection mechanism that identifies the most generalizable abstraction from a set of candidate summaries. This summary is prepended to inputs from the held-out test split at inference-time which enables the model to generalize from a small set of explanations without updating model weights or performing multi-turn interaction.

Notably, we find that only 4-11 verified explanations are sufficient to induce improvements across diverse benchmarks and model scales. Across CounterBench \cite{chen2025counterbench}, GSM8K \cite{cobbe2021training}, and ReasonIF \cite{kwon2025reasonif}, FLEx yields consistent gains on both the Gemma \cite{team2025gemma} and Qwen \cite{yang2025qwen3} model families, with some improvements reaching over 25 percentage points and reductions in residual errors surpassing 80\%. These improvements hold across model scales from 0.5B to 72B parameters.

This behavior parallels aspects of human learning, where targeted feedback on a small number of mistakes can help expose underlying principles and support generalization. Our results suggest that modern LLMs already encode reasoning behaviors that can be elicited through a small number of high-quality, error-corrective explanations. More broadly, these findings indicate that inference-time guidance in natural language can, in certain settings, offer a lightweight alternative to large-scale annotation or parameter fine-tuning.

Our contributions are as follows:

\begin{itemize}
    \item We introduce FLEx, a lightweight framework that improves frozen LLMs using only a small number of verified natural language explanations.
    \item We introduce a clustering-based error selection procedure and a summary selection method that identify representative failure modes and distill corrective explanations into a compact prompt prefix that generalizes across inputs.
    \item We demonstrate the effectiveness of FLEx on CounterBench, GSM8K, and ReasonIF, achieving consistent gains across multiple LLM architectures and parameter scales.

\end{itemize}

\section{Related Work}

% \paragraph{Explanations as Supervision}
% Several approaches train or fine-tune language models using natural language explanations. \citet{zelikman2022star} propose STaR, an iterative method where the model generates rationales, filters for correct answers, and fine-tunes on the filtered data. This improves reasoning performance without requiring large annotated datasets. \citet{chen2024learning} introduce ILF, an imitation learning approach that uses human-written feedback explaining how to improve flawed model outputs. They show that incorporating feedback alongside standard examples improves performance across tasks. These methods demonstrate that explanations can serve as effective supervision but require model updates and larger-scale data.

We focus on three relevant strategies for improving model behavior: self-refinement, retrieval-augmented generation, and test-time scaling.

\paragraph{Self-Refinement.}
Several approaches focus on iterative refinement using natural language explanations at inference-time. In ReFLExion \cite{shinn2023reFLExion}, an LLM reflects on its mistakes and uses generated critiques to improve future responses without weight updates. This method improves decision-making and coding tasks through multi-turn self-correction. Related approaches \cite{madaan2023self, huang2022large, ouyang2025reasoningbank, richardson2023syndicom} show that self-critiquing and retrying can boost accuracy on math and reasoning tasks. These methods typically rely on multiple rounds of generation. FLEx differs by using a fixed set of verified explanations to guide inference in a single forward pass.

\paragraph{Retrieval-augmented generation.}
Explanations have also been used to retrieve relevant examples. \citet{wu2024discret} treat explanations as structured queries that retrieve similar prior cases via embedding or rule-based search. Their method, DISCRET, enables analogy-based reasoning by identifying instances with similar underlying logic. While effective, these approaches require a retrieval mechanism and external database. FLEx instead summarizes a small set of explanations into a single prompt prefix, avoiding retrieval while still promoting generalization.

\paragraph{Test-time scaling.}
A complementary line of work improves model performance by allocating additional computation at inference-time rather than modifying model parameters \cite{wang2022self, yao2023tree, besta2024graph, zhou2022least}. One instance of this idea is self-consistency \cite{wang2022self}, which samples multiple reasoning traces for the same input and selects a final answer based on agreement across samples. While test-time scaling improves robustness to stochastic errors, it operates independently per input and does not accumulate information across examples, limiting its ability to correct systematic failure modes. In contrast, FLEx explicitly targets systematic, recurring errors by generalizing
corrective explanations across inputs.

\paragraph{Comparison to FLEx.}
FLEx enables generalization from a small number of verified explanations without fine-tuning or multi-turn interaction. In contrast to self-refinement that relies on repeated inference, retrieval-augmented approaches that require maintaining external databases, and test-time scaling techniques that improve robustness per input without correcting recurring failures, FLEx distills corrective explanations into a single prompt-level abstraction applied in one forward pass. This positions FLEx as a lightweight alternative between prompting and model fine-tuning.

\section{Method}

\subsection{Preliminaries} \label{sec:feedback_preliminaries}

We consider a standard causal language modeling setup. Let $x$ denote an input, $y$ its gold target, and let a frozen language model $M$ produce a response $r = M(x)$. A task-specific binary scoring function $S(r,y) \in \{0,1\}$ indicates whether the model output is correct.

Our training split provides inputs for which we can collect explanations. We denote by
\[
\mathcal{E} = \{(x,r,y) \mid S(r,y) = 0\}
\]
the set of all model errors on the training data, where each element consists of the input $x$, the model's incorrect response $r$, and the ground-truth label $y$. These errors serve as candidates for annotation.

Our objective is to use only a small subset of these errors to construct generalizable inference-time improvements. Given a small annotated set of explanations $\mathcal{F}$ ($|\mathcal{F}| \ll|\mathcal{E}|$), the goal is to prepend a learned summary $s^\star$ to any future input $x$ so that the frozen model's prediction $\hat y = M([s^\star; x])$ is more accurate, without modifying model parameters.

% \begin{algorithm}[h]
% \caption{\textsc{FLEx}}
% \label{alg:FLEx}
% \begin{algorithmic}
%     \State \textbf{1. Run baseline inference.}
%     \State Compute error set $\mathcal E = \{(x,r) : M(x)=r,\ S(r,y)=0\}$.

%     \State \textbf{2. Cluster errors.}
%     \State Embed each $z=[x;r]$ with $\phi(z)$.
%     \State Select $k^*$ using the WCSS elbow.
%     \State Run $k^*$-means; select the 5 points nearest each centroid.

%     \State \textbf{3. Collect verified feedback.}
%     \For{each selected error $(x,r)$}
%         \State Write explanation $f$.
%         \While{$M([x;r;f])$ incorrect}
%             \State Rewrite $f$ \textbf{or} move to next sample in cluster.
%         \EndWhile
%         \State Add $(x,r,f)$ to $\mathcal F$.
%     \EndFor

%     \State \textbf{4. Generate candidate summaries.}
%     \State Use GPT-4o to sample $L$ summaries conditioned on $\mathcal F$.

%     \State \textbf{5. Select best summary.}
%     \State For each summary $s^{(\ell)}$, compute
%     \[
%        J(\ell)=\sum_i w_i\,\cos\big(\Delta_f(x_i),\,\Delta_s^{(\ell)}(x_i)\big)
%     \]
%     \State Return $s^*=\arg\max_\ell J(\ell)$.
% \end{algorithmic}
% \end{algorithm}

% \subsection{Preliminaries}
% Let $x$ denote an input, $y$ a gold label, and $r=M(x)$ the model response.  A task-specific binary success function $S(r,y)\in\{0,1\}$ indicates correctness.  Denote the set of \emph{errors} on the training split as $\mathcal E\!:=\!\{(x,r)\mid S(r,y)=0\}$.

\subsection{Error clustering and sample selection} The first step is to select which error samples to annotate with explanations. Our goal is to obtain representative examples across distinct error types so that the resulting explanations generalize effectively. To achieve this, we embed each concatenated pair $[x; r]$ using an encoding function $\phi : \mathcal X \to \mathbb R^d$, instantiated as the final-layer hidden state of the last token of the concatenated sequence $[x;r]$, extracted from the same frozen model $M$ used at inference-time. We run $k$-means clustering \cite{mcqueen1967some} over these embeddings to partition the error set into clusters ${C_1, \dots, C_k}$. 

% To determine the number of clusters, we use the elbow method \cite{thorndike1953belongs}: for each candidate 
% value of $k$, we compute the within-cluster sum of squared distances (WCSS),
% \[
% \mathrm{WCSS}(k) = \sum_{j=1}^{k} \sum_{z_i \in C_j} \lVert z_i - \mu_j \rVert_2^2,
% \]
% where $\mu_j$ is the centroid of cluster $C_j$. We select the `elbow' - the point 
% beyond which additional clusters yield diminishing returns.

% For each cluster, we select the sample closest to the centroid as the primary 
% representative and include the next four nearest samples, yielding five candidates 
% in total. These secondary examples serve as backups when the primary instance is 
% unclear or difficult to annotate.

To determine the number of clusters, we use a standard elbow-based criterion that selects the point beyond which increasing $k$ yields diminishing reductions in within-cluster variance (Appendix~\ref{app:clustering_details}). Once the clusters are formed, we select one representative error per cluster-the instance closest to the cluster centroid-as the primary candidate for annotation. We additionally retain the next four nearest neighbors within each cluster as backups in case the primary example is unclear or difficult to annotate.

\subsection{Explanations and Verification}
\begin{figure}[t]
    \centering
    \includegraphics[width=.95\columnwidth]{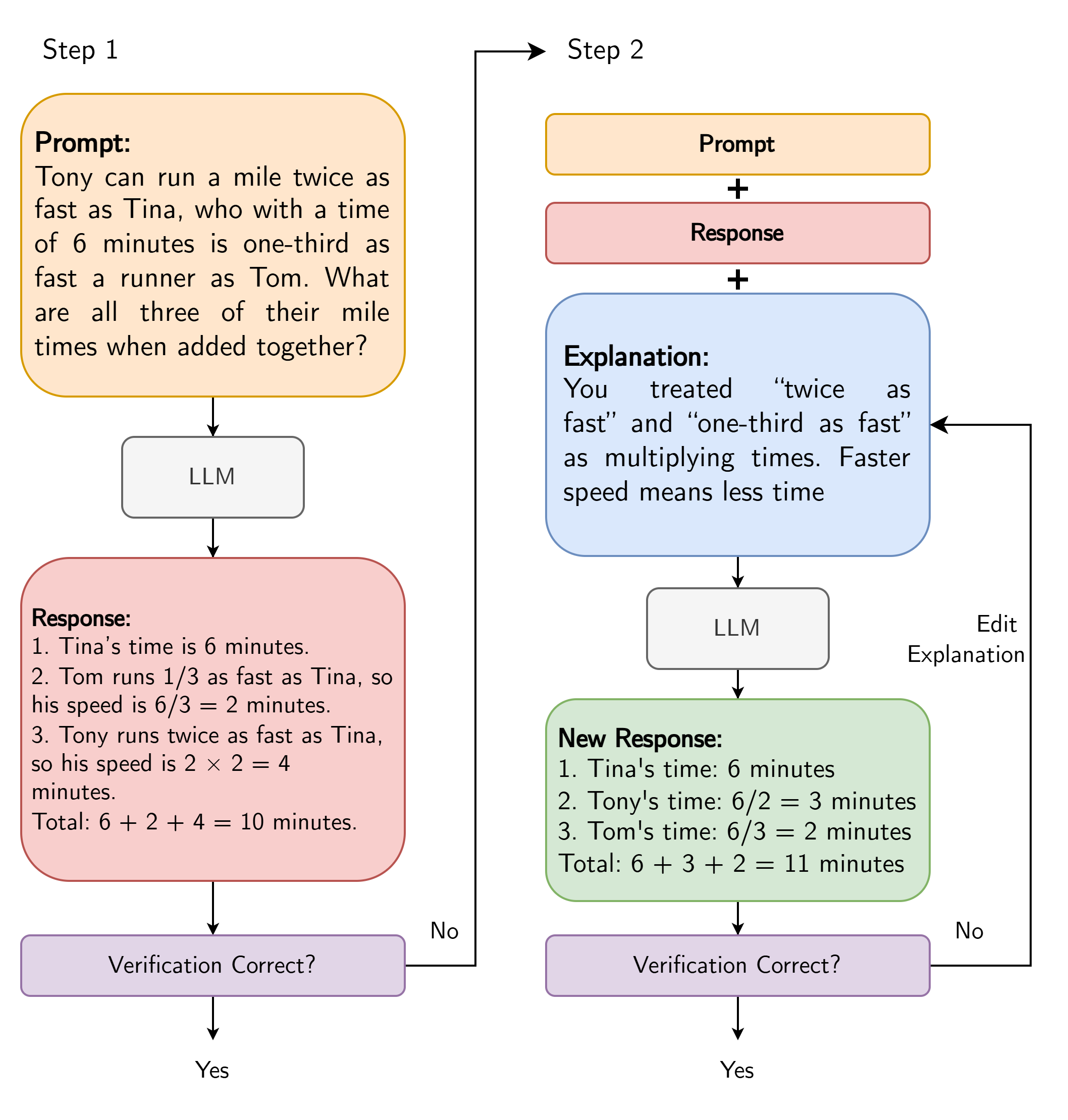}
    % \caption{Two-step explanation verification pipeline used in FLEx. The LLM first produces an incorrect response (Step 1). An annotator then provides a corrective explanation, which is appended to the original prompt and response. The LLM re-evaluates this combined input (Step 2), and if the output is still incorrect, the feedback is edited until the example is successfully corrected.}
    \caption{Example of explanation verification}
    \label{fig:feedbackverification}
\end{figure}

For each selected input-erroneous response pair $(x,r)$, an annotator writes an explanation $f$ expected to guide $M$ to the correct answer when prepended: $S\bigl(M([x;r;f]),y\bigr)=1$. Annotators iterated on explanations until this condition was satisfied. If the annotator was unable to obtain a correct model response for a given sample after several attempts, we discarded that sample and moved to the next closest example to the cluster mean. This verification process results in $\mathcal F=\{(x_i,r_i,f_i)\}_{i=1}^{k}$. An example of this iterative verification process is shown in Figure \ref{fig:feedbackverification}.

\subsection{Summarization}
To generalize from the annotated explanations, we take inspiration from \citet{richardson2023integrating} and distill the verified explanations into a concise, reusable summary that replaces the individual annotations at inference-time. We use a separate, off-the-shelf chat LLM as an external summarizer to avoid entangling the summarization process with the behavior of the target models under evaluation. In particular, we employ GPT-4o mini \cite{hurst2024gpt}, which we refer to as the summarizer $S_{\text{sum}}$, due to its strong abstractive summarization performance and reliable instruction following.

Given the verified explanation set $\mathcal F$ for a particular model and dataset, the summarizer generates a collection of candidate summaries $\{s^{(\ell)}\}_{\ell=1}^{L}$. To encourage diversity, we use five distinct summary prompts: three written by the authors and two generated by the summarizer itself (Table~\ref{tab:summary_prompts}). For each prompt, we condition the summarizer on the prompt together with a JSON representation of $\mathcal F$, and sample ten summaries using a temperature of $T=1.0$. This results in a total of $L=50$ candidate summaries per model--dataset pair.

% \subsection{Summarization}
% To generalize from the annotated explanations, we take inspiration from \citet{richardson2023integrating} and distill the verified feedback into a concise, reusable summary that replaces the individual annotations at inference-time. We use a separate, off-the-shelf chat LLM as an \emph{external summarizer} to avoid entangling the summarization process with the behavior of the target models under evaluation. In particular, We employ GPT-4o mini \cite{hurst2024gpt} for this purpose, which we refer to as the \textit{summarizer}, $S_{\text{sum}}$, which produces $L$ candidate summaries $\{s^{(\ell)}\}_{\ell=1}^{L}$ conditioned on $\mathcal F$. To generate the summaries, we employ 5 summary prompts: 3 written by the authors and 2 written by the summarizer model itself (Table \ref{tab:summary_prompts}). For each summary prompt, we prompt the summarizer model with the summary prompt and attach a JSON file containing $\mathcal F$. We use a temperature of $T=1.0$ to generate 10 different summaries for each summary prompt for each summarizer model, resulting in a total of $L=50$ different summaries for each dataset.

\subsection{Summary Selection}
Each summary is concatenated with inputs to form a new prompt prefix. Inspired by prior work that analyzes and steers model behavior via internal representation shifts \citep{turner2024activation}, we evaluate each summary by comparing the change in the representation induced by the explanations to the change induced by the summary.  Let $\phi(z) \in \mathbb{R}^d$ denote the final-layer hidden state of the last input token of $z$, extracted from the same frozen model $M$ used at inference-time.  For each training example $(x_i, r_i, f_i)$ and candidate summary $s^{(\ell)}$, we define the explanation-induced and summary-induced deltas as
\begin{align}
    \Delta_f(x_i) &= \phi([x_i; r_i; f_i]) - \phi([x_i; r_i]), \\
    \Delta_s^{(\ell)}(x_i) &= \phi([x_i; r_i; s^{(\ell)}]) - \phi([x_i; r_i]).
\end{align}

Because one representative example is selected from each cluster but clusters vary in size, not all explanation examples should contribute equally. Let $C(x_i)$ denote the cluster from which example $(x_i, r_i, f_i)$ was selected. Larger clusters correspond to more frequent error modes and should therefore receive proportionally higher weight. We define a normalized cluster weight
\[
    w_i = \frac{|C(x_i)|}{\sum_{j=1}^{k} |C(x_j)|},
\]
where $|C(x_i)|$ is the size of the cluster associated with example $i$. We score each summary using a cluster-weighted average cosine similarity:
\begin{equation}
    J(\ell) = \sum_{i=1}^{k} 
    w_i \, \cos(\Delta_f(x_i),\, \Delta_s^{(\ell)}(x_i)).
\end{equation}
We first select the optimal summary index
\[
\ell^{\star} = \arg\max_{\ell} J(\ell),
\]
and define the final summary as $s^{\star} = s^{(\ell^{\star})}$. At test-time we prepend $s^{\star}$ to the prompt, i.e., $\hat y = M([s^{\star}; x])$, and do not update any model weights.

\section{Experiments}\label{sec:experiments}

\begin{table*}[t]
\centering
\resizebox{\textwidth}{!}{
\begin{tabular}{
    l
    c c c c c  % CounterBench
    c         % spacer
    c c c c c   % GSM8K
    c         % spacer
    c c c c c % ReasonIF
}
\toprule
& \multicolumn{5}{c}{\textbf{CounterBench}} &
  \multicolumn{1}{@{}c@{}}{\hspace{0.9em}} &
  \multicolumn{5}{c}{\textbf{GSM8K}} &
  \multicolumn{1}{@{}c@{}}{\hspace{0.9em}} &
  \multicolumn{5}{c}{\textbf{ReasonIF}} \\
\cmidrule(lr){2-6}
\cmidrule(lr){8-12}
\cmidrule(lr){14-18}
\textbf{Model}
& CoT & SR & RAG & SC & FLEx
& \multicolumn{1}{@{}c@{}}{}
& CoT & SR & RAG & SC & FLEx
& \multicolumn{1}{@{}c@{}}{}
& CoT & SR & RAG & SC & FLEx \\
\midrule

\rowcolor{lightlightgray} 
\multicolumn{18}{c}{\textbf{\textit{Gemma-3 Instruct}}} \\

Gemma-1B &
49.8 & 49.7 & 48.5 & 51.1 & \textbf{51.6} &
\multicolumn{1}{@{}c@{}}{} &
45.6 & 44.8 & 38.0 & \textbf{53.2} & 46.0 &
\multicolumn{1}{@{}c@{}}{} &
35.3 & 38.0 & 46.3 & 36.3 & \textbf{50.0} \\

Gemma-4B &
64.9 & 65.4 & 69.7 & 68.5 & \textbf{74.3} &
\multicolumn{1}{@{}c@{}}{} &
85.4 & 84.8 & 85.2 & \textbf{87.7} & 86.1 &
\multicolumn{1}{@{}c@{}}{} &
50.0 & 47.3 & 58.0 & 46.7 & \textbf{66.0} \\

Gemma-12B &
68.7 & 66.7 & 70.4 & 72.1 & \textbf{84.9} &
\multicolumn{1}{@{}c@{}}{} &
93.4 & 93.1 & 92.9 & \textbf{94.0} & 93.8 &
\multicolumn{1}{@{}c@{}}{} &
64.3 & 58.0 & 71.0 & 63.3 & \textbf{74.0} \\

Gemma-27B &
76.0 & 74.6 & 75.9 & 78.2 & \textbf{80.7} &
\multicolumn{1}{@{}c@{}}{} &
94.2 & 94.7 & 94.5 & 95.1 & \textbf{95.5} &
\multicolumn{1}{@{}c@{}}{} &
75.0 & 68.7 & \textbf{77.3} & 73.7 & 76.7 \\

\midrule
\rowcolor{lightlightgray} 
\multicolumn{18}{c}{\textbf{\textit{Qwen-2.5 Instruct}}} \\

Qwen-0.5B &
21.3 & 21.8 & \textbf{43.4} & 38.0 & 42.6 &
\multicolumn{1}{@{}c@{}}{} &
24.8 & 29.0 & \textbf{40.2} & 34.3 & 24.9 &
\multicolumn{1}{@{}c@{}}{} &
28.3 & 35.7 & \textbf{45.0} & 29.0 & 35.3 \\

Qwen-1.5B &
39.3 & 47.0 & \textbf{49.7} & 47.8 & 47.8 &
\multicolumn{1}{@{}c@{}}{} &
50.3 & 51.8 & \textbf{69.4} & 59.0 & 53.4 &
\multicolumn{1}{@{}c@{}}{} &
30.3 & 36.3 & \textbf{48.0} & 31.3 & 35.0 \\

Qwen-3B &
50.5 & 53.3 & \textbf{61.3} & 55.2 & 58.1 &
\multicolumn{1}{@{}c@{}}{} &
83.9 & 84.0 & 85.6 & \textbf{87.6} & 84.8 &
\multicolumn{1}{@{}c@{}}{} &
36.3 & \textbf{44.7} & 38.7 & 39.3 & 42.3 \\

Qwen-7B &
69.9 & 68.8 & 68.5 & 69.0 & \textbf{78.0} &
\multicolumn{1}{@{}c@{}}{} &
90.2 & 88.2 & 90.7 & \textbf{93.0} & 91.0 &
\multicolumn{1}{@{}c@{}}{} &
46.3 & 56.7 & 66.7 & 47.3 & \textbf{67.7} \\

Qwen-14B &
74.0 & 74.7 & 78.3 & 75.8 & \textbf{80.6} &
\multicolumn{1}{@{}c@{}}{} &
91.6 & 91.7 & 94.4 & \textbf{95.5} & 94.8 &
\multicolumn{1}{@{}c@{}}{} &
71.3 & 70.0 & 75.3 & 72.3 & \textbf{95.0} \\

Qwen-32B &
79.7 & 80.3 & \textbf{85.9} & 80.2 & 84.6 &
\multicolumn{1}{@{}c@{}}{} &
92.1 & 94.1 & 94.8 & 92.5 & \textbf{96.4} &
\multicolumn{1}{@{}c@{}}{} &
78.3 & 75.0 & 79.7 & 79.0 & \textbf{96.3} \\

Qwen-72B &
82.8 & 81.3 & 86.5 & 85.2 & \textbf{88.9} &
\multicolumn{1}{@{}c@{}}{} &
93.9 & 94.8 & 95.1 & 93.6 & \textbf{95.7} &
\multicolumn{1}{@{}c@{}}{} &
87.0 & 85.3 & 83.7 & 83.0 & \textbf{93.7} \\

\midrule
\textbf{Avg. CoT $\Delta$} &
0.0 & $\uparrow$ 0.6 & $\uparrow$ 5.6 & $\uparrow$ 4.0 & \textbf{$\uparrow$ 8.7} &
& 0.0 & $\uparrow$ 0.5 & $\uparrow$ 3.2 & \textbf{$\uparrow$ 3.6} & $\uparrow$ 1.5 &
& 0.0 & $\uparrow$ 1.2 & $\uparrow$ 8.1 & $\uparrow$ 0.1 & \textbf{$\uparrow$ 11.8} \\

\bottomrule
\end{tabular}
}
\caption{
Unified evaluation across CounterBench, GSM8K, and ReasonIF.
FLEx outperforms Zero-shot Chain-of-Thought (CoT), Self Refine (SR) \citep{madaan2023self, bai2022constitutional, shinn2023reFLExion}, Retrieval-Augmented Generation (RAG) \cite{wu2024discret}, and Self-Consistency (SC) \cite{wang2022self}, across most settings.}
\label{tab:main_results}
\end{table*}

We evaluate FLEx on three diverse benchmarks that span robustness, logical reasoning, and instruction following. 

\subsection{Datasets}

\paragraph{CounterBench.}
CounterBench \cite{chen2025counterbench} measures counterfactual robustness by introducing minimally perturbed instructions that require the model not to follow misleading, inconsistent, or self-contradictory cues. Models are evaluated using exact constraint adherence under counterfactual shifts.

\paragraph{GSM8K.}
GSM8K \cite{cobbe2021training} is an industry standard benchmark for grade-school mathematics word problems requiring multi-step numerical reasoning. We evaluate using the standard exact-match accuracy on the final answer.

\paragraph{ReasonIF.}
ReasonIF \cite{kwon2025reasonif} is a benchmark for evaluating the degree to which large reasoning models follow verifiable instructions within their reasoning traces. Each example pairs a question with a single constraint, and performance is measured using the instruction-following score (IFS) computed by an automatic checker. Since no training split was present, we synthesized one following the procedure described in Appendix~\ref{sec:reasonif}.

\subsection{Baselines}

We compare FLEx against a set of training-free baselines that improve reasoning performance without modifying model parameters. For annotation purposes, we run chain-of-thought (CoT) on the training split to expose the model's reasoning on its errors, enabling annotators to diagnose failure modes and write targeted corrective explanations. These training rationales are not used at test-time.

% \paragraph{Zero-shot Chain-of-Thought (CoT).}
% Our primary baseline is standard zero-shot CoT prompting on the test split. Model outputs are evaluated using each dataset's official scoring procedure.

\paragraph{Chain-of-Thought (CoT).}
Our main baseline is standard zero-shot CoT prompting on the test split.

\paragraph{Self-Refine (SR).}
We implement a single-step critique-and-revise baseline inspired by prior work \citep{madaan2023self, bai2022constitutional, shinn2023reFLExion}. On the test split, the model first produces a zero-shot CoT solution, then generates a natural-language critique of that solution, and finally outputs a revised response conditioned on both the original reasoning and the critique.

\paragraph{Retrieval-augmented generation (RAG).}
We include a RAG baseline, inspired by prior work \cite{wu2024discret, lewis2020retrieval}. We first run zero-shot CoT on the training split and collect only correct model responses. These correct training examples are indexed using dense embeddings, and at test-time, for each input we retrieve the most similar solved example and prepend it to the prompt as an in-context demonstration.

\paragraph{Self-Consistency (SC).}
Following \citet{wang2022self}, we sample $n$ CoT trajectories for each test input and select the final answer by majority vote. This improves robustness by marginalizing over diverse reasoning paths. For ReasonIF, we additionally retain the first reasoning
trace corresponding to the most frequently selected answer, and use this trace for instruction-following evaluation.

\subsection{Results}

% Table~\ref{tab:main_results} reports the results on CounterBench, GSM8K, and ReasonIF for the Gemma-3 and Qwen-2.5 model families. We compare five strategies: (i) standard zero-shot chain-of-thought prompting (CoT), (ii) a single-step self-refinement procedure (SR), (iii) a rag pipeline (RAG), (iv) self-consistency decoding with $n=5$ samples (SC) and (v) our proposed method, FLEx. Across all benchmarks and model scales, FLEx outperforms or matches the strongest baseline in the majority of settings, and achieves the best average performance overall.

Table~\ref{tab:main_results} reports results for the Gemma-3 and Qwen-2.5 model families. On CounterBench, FLEx achieves the strongest average gains (+8.7) and performs best for medium and large models, while RAG is competitive for smaller Qwen models, reflecting the benefit of retrieval when base reasoning is limited. On GSM8K, improvements are more modest and distributed across methods: SC and RAG often achieve the strongest results for smaller models, whereas FLEx provides consistent gains that increase with model scale. On ReasonIF, FLEx outperforms all baselines across most models, yielding the largest average improvement (+11.8). Overall, FLEx outperforms the strongest baseline in most settings and achieves the highest average improvement across benchmarks.

\paragraph{Error-rate reduction.}
Although accuracy gains in larger models appear small, many already operate in the 80--95\% accuracy range, making absolute accuracy differences less informative. In this regime, we additionally report error-rate reduction (ERR), defined as the percentage decrease in the residual errors relative to zero-shot CoT. This metric captures how effectively a method eliminates hard-to-fix failures. Table~\ref{tab:error_reduction} reports ERR for all models and datasets. On average, FLEx reduces residual errors by 24.6\% on CounterBench, 16.0\% on GSM8K, and 33.8\% on ReasonIF. In several settings, FLEx eliminates more than 80\% of the errors made by zero-shot CoT, demonstrating that targeted, prompt-level corrections can mitigate systematic failure modes.

\begin{table}[h]
\centering
\small
\setlength{\tabcolsep}{6pt}
\begin{tabular}{lccc}
\toprule
\textbf{Model} & \textbf{CounterBench} & \textbf{GSM8K} & \textbf{ReasonIF} \\
\midrule
Gemma-1B   & 3.59  & 0.70  & 22.08 \\
Gemma-4B   & 26.78 & 4.17  & 32.00 \\
Gemma-12B  & 51.76 & 6.25  & 27.10 \\
Gemma-27B  & 19.58 & 22.08 & 6.70  \\
\midrule
Qwen-0.5B  & 27.06 & 0.10  & 9.77  \\
Qwen-1.5B  & 14.00 & 6.25  & 6.70  \\
Qwen-3B    & 15.35 & 5.63  & 9.42  \\
Qwen-7B    & 26.91 & 7.75  & 39.75 \\
Qwen-14B   & 25.38 & 37.84 & 82.56 \\
Qwen-32B   & 24.14 & 53.85 & 83.08 \\
Qwen-72B   & 35.47 & 29.63 & 51.28 \\
\midrule
\textbf{Average} & 24.55 & 16.01 & 33.78 \\
\bottomrule
\end{tabular}
\caption{Error-rate reduction (ERR, \%) achieved by FLEx relative to zero-shot CoT, computed as the percentage reduction in residual errors.}
\label{tab:error_reduction}
\end{table}

When comparing methods using error-rate reduction rather than absolute accuracy, a trend emerges: larger models often benefit the most from FLEx. Although high-capacity models start from stronger baselines, FLEx consistently removes a larger fraction of their remaining errors, indicating that prompt-level guidance can meaningfully improve performance even at scale.

\section{Analysis}

FLEx consists of three core components: (i) sample selection, (ii) explanation and verification, and (iii) summarization and summary selection. In this section, we present targeted ablations to analyze the contribution of each component. Unless otherwise stated, all reported numbers in this section are average accuracy deltas computed relative to the zero-shot CoT baseline and averaged across all evaluated models, denoted $\Delta\text{Acc}$ with ($\uparrow$) indicating improvement and ($\downarrow$) indicating degradation. Full per-model results are provided in Appendix~\ref{sec:ablations}.

\paragraph{Effect of error selection.}
We begin by evaluating the importance of the error selection procedure. FLEx uses $k$-means clustering to identify diverse error modes, ensuring that the selected error samples represent distinct failure patterns. We compare clustering-based selection with random sampling, holding the number of selected errors $k$ fixed in both cases, and analyze their influence on overall performance. As shown in Table~\ref{tab:error_selection}, $k$-means sampling consistently outperforms random selection across all datasets.

\begin{table}[h]
\centering
\begin{tabular}{lcc}
\toprule
\textbf{Dataset} & \textbf{$k$-means} & \textbf{Random}  \\
\midrule
CounterBench    & \textbf{$\uparrow$ 8.66\phantom{0}} & $\uparrow$ 3.46  \\
GSM8K           & \textbf{$\uparrow$ 1.55\phantom{0}} & $\downarrow$ 5.65 \\
ReasonIF        & \textbf{$\uparrow$ 11.99} & $\uparrow$ 2.58  \\
\midrule
\textbf{Average} & \textbf{$\uparrow$ 8.95\phantom{0}} & $\uparrow$ 0.13  \\
\bottomrule
\end{tabular}
\caption{$\Delta\text{Acc}$ vs error selection strategies. Values indicate average accuracy change (relative to zero-shot CoT) averaged across models.}
\label{tab:error_selection}
\end{table}

Some datasets, such as ReasonIF, include predefined task types. In ReasonIF, each example is assigned a structural or stylistic constraint the model's reasoning must follow. To assess whether $k$-means provides additional value beyond this task structure, we construct an alternative baseline that selects one error at random from each of the six task types provided by the dataset authors.

\begin{figure}[h]
\includegraphics[width=1\linewidth]{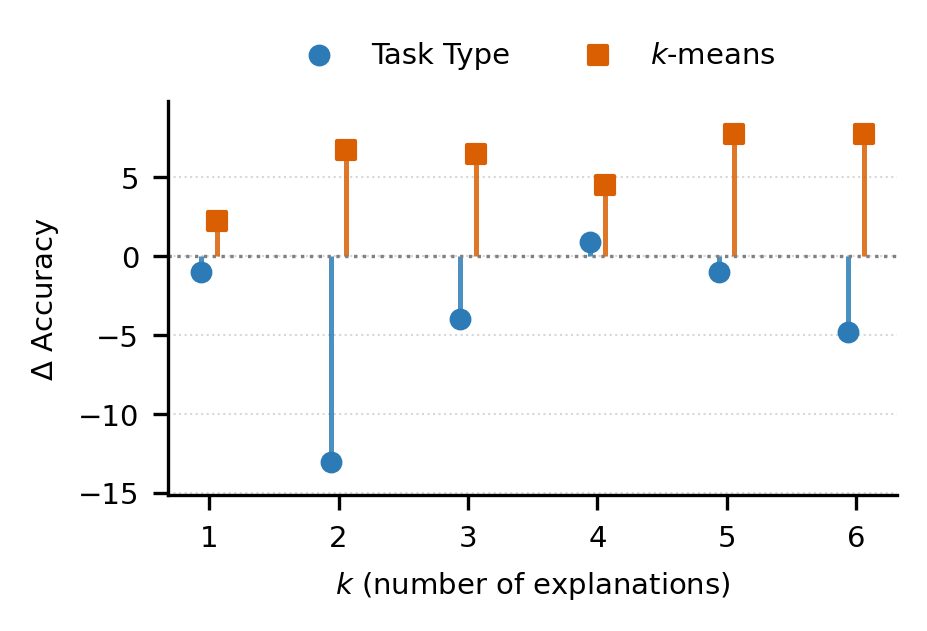}
\caption{$\Delta\text{Acc}$ for Task-Type vs $k$-means clustering as a function of $k \in \{1,2,3,4,5,6\}$ on ReasonIF.}
\label{fig:k_curve}
\end{figure}
Figure~\ref{fig:k_curve} shows performance as we vary the number of selected errors $k$, starting from the most frequent failure mode and progressively adding additional modes. Across all values of $k$, $k$-means consistently outperforms task-type selection. While task-type selection ensures coarse coverage of reasoning skills, $k$-means captures finer-grained distinctions between failure modes, leading to more effective explanatory examples.

\paragraph{Impact of explanation quality.}
We next evaluate the importance of the explanations themselves. FLEx relies on verified, error-corrective explanations: each annotation is iteratively refined until appending it to the model's original response induces the correct prediction. To assess whether this verification procedure is necessary, we compare three conditions: (i) \emph{verified} explanations produced through this iterative correction process; (ii) \emph{unverified} explanations generated automatically by GPT-4o mini and used without verification; and (iii) \emph{solutions only}, where the summarizer is given the correct solution in place of an explanation.

\begin{table}[h]
\centering
\resizebox{\columnwidth}{!}{
\begin{tabular}{lccc}
\toprule
\textbf{Dataset} & \textbf{Verified} & \textbf{Unverified} & \textbf{Solution} \\
\midrule
CounterBench    & \textbf{$\uparrow$ 8.66\phantom{0}}  & $\downarrow$ 2.49 & $\uparrow$ 0.34 \\
GSM8K           & \textbf{$\uparrow$ 1.55\phantom{0}} & $\downarrow$ 8.31 & $\downarrow$ 2.11 \\
ReasonIF        & \textbf{$\uparrow$ 11.99}  & $\downarrow$ 0.50 & $\downarrow$ 0.25 \\
\midrule
\textbf{Average} & \textbf{$\uparrow$ 8.95\phantom{0}} & $\downarrow$ 3.77 & $\downarrow$ 0.67 \\
\bottomrule
\end{tabular}
}
\caption{$\Delta\text{Acc}$ vs the quality of explanatory explanations. Values indicate average accuracy change relative to zero-shot CoT, averaged across evaluated models.}
\label{tab:feedback_quality}
\end{table}

As shown in Table~\ref{tab:feedback_quality}, verified explanations consistently yield the largest improvements, while unverified LLM-generated explanations are unreliable and often harmful. In contrast, solution-only summaries-without explicit error explanations-provide little or no benefit. These results indicate that FLEx's gains arise from high-quality, error-diagnostic explanations rather than exposure to correct answers alone.

% As shown in Table~\ref{tab:feedback_quality}, explanation quality strongly affects performance. Verified explanations consistently yield positive gains across all datasets, while unverified LLM-generated explanations are unreliable and often harmful. In contrast, solution-only summaries—without explicit error explanations—provide little or no benefit. These results indicate that FLEx’s gains arise from high-quality, error-diagnostic explanations rather than exposure to correct answers alone.

\paragraph{Importance of summarization.}
We examine whether abstracting explanations improves generalization. Table~\ref{tab:summarization} shows that summarized explanations consistently outperform raw concatenation of explanations, which often degrades performance. This demonstrates that distilling explanations into concise, general principles is essential for effective generalization.

\begin{table}[h]
\centering
\resizebox{\columnwidth}{!}{
\begin{tabular}{lcc}
\toprule
\textbf{Dataset} & \textbf{Summary} & \textbf{Raw Expl.} \\
\midrule
CounterBench    & \textbf{$\uparrow$ 8.66\phantom{0}} & $\uparrow$ 2.22 \\
GSM8K           & \textbf{$\uparrow$ 1.55\phantom{0}} & $\downarrow$ 5.96 \\
ReasonIF        & \textbf{$\uparrow$ 11.99}  & $\downarrow$ 2.34 \\
\midrule
\textbf{Average} & \textbf{$\uparrow$ 8.95\phantom{0}} & $\downarrow$ 2.04 \\
\bottomrule
\end{tabular}
}
\caption{$\Delta\text{Acc}$ vs summarization method. Values indicate average accuracy change relative to zero-shot CoT, averaged across evaluated models.}
\label{tab:summarization}
\end{table}

\paragraph{Effectiveness of summary selection.}
We evaluate the contribution of FLEx's summary-selection mechanism, which ranks candidate summaries using the $\Delta$-embedding similarity score. If the score is meaningful, higher-ranked summaries should yield larger downstream gains. As shown in Table~\ref{tab:summary_rank}, the highest-scoring summaries consistently outperform median-ranked ones, while the lowest-ranked summaries often degrade performance. This confirms that the $\Delta$-embedding metric reliably identifies generalizable summaries.

\begin{table}[h]
\centering
\resizebox{\columnwidth}{!}{
\begin{tabular}{lccc}
\toprule
\textbf{Dataset} & \textbf{Best} & \textbf{Median} & \textbf{Worst} \\
\midrule
CounterBench    & \textbf{$\uparrow$ 8.66\phantom{0}} & $\uparrow$ 2.80 & $\downarrow$ 0.14\\
GSM8K           & \textbf{$\uparrow$ 1.55\phantom{0}} & $\downarrow$ 0.56 & $\downarrow$ 2.14 \\
ReasonIF        & \textbf{$\uparrow$ 11.99} & $\uparrow$ 5.79 & $\uparrow$ 2.09 \\
\midrule
\textbf{Average} & \textbf{$\uparrow$ 8.95\phantom{0}} & $\uparrow$ 2.68 & $\downarrow$ 0.06\\
\bottomrule
\end{tabular}
}
\caption{$\Delta\text{Acc}$ vs summary ranking. Values indicate average accuracy change relative to zero-shot CoT, averaged across evaluated models.}
\label{tab:summary_rank}
\end{table}

We further observe that summary selection is more critical for smaller models: low-ranked summaries can substantially degrade performance, whereas larger models are comparatively more robust to summary quality, with the magnitude of this effect varying by dataset (Table~\ref{tab:ablation_summary_rank_full}).

\begin{table}[h]
\centering
\resizebox{\columnwidth}{!}{
\begin{tabular}{lcc}
\toprule
\textbf{Dataset} & \textbf{Weighted} & \textbf{Unweighted} \\
\midrule
CounterBench    & \textbf{$\uparrow$ 8.66\phantom{0}} & $\uparrow$ 3.54 \\
GSM8K           & \textbf{$\uparrow$ 1.55\phantom{0}} & $\downarrow$ 0.01 \\
ReasonIF        & \textbf{$\uparrow$ 11.99} & $\uparrow$ 7.02 \\
\midrule
\textbf{Average} & \textbf{$\uparrow$ 8.95\phantom{0}} & $\uparrow$ 3.52 \\
\bottomrule
\end{tabular}
}
\caption{$\Delta\text{Acc}$ vs weighted summary scoring. Values indicate average accuracy change relative to zero-shot CoT, averaged across evaluated models.}
\label{tab:weighted_unweighted}
\end{table}

In addition to ranking summaries, we test whether the scoring function should account for the prevalence of different failure modes. We compare the original unweighted score to a cluster-weighted variant that assigns higher weight to more frequent error clusters. As shown in Table~\ref{tab:weighted_unweighted}, cluster-weighted scoring yields consistent improvements, indicating that emphasizing dominant error patterns improves summary selection stability.

\paragraph{Compatibility with existing methods.}
FLEx functions as a prompt-level adapter and is complementary. It can be applied on top of existing pipelines such as self-refine (SR), retrieval-augmented generation (RAG), and self-consistency (SC) without changing their implementations. Table~\ref{tab:flex_method_delta} shows that augmenting these methods with FLEx consistently improves accuracy relative to the corresponding method alone. Full results are reported in Table~\ref{tab:flex_plus_others}, where we find that in many cases, combining FLEx with an existing method also outperforms FLEx alone.

% As shown in Table~\ref{tab:flex_plus_others} in the appendix, combining FLEx with these methods often yields additional gains over either method alone, indicating that FLEx provides orthogonal benefits.
\begin{table}[h]
\centering
\resizebox{\columnwidth}{!}{
\begin{tabular}{lccc}
\toprule
\textbf{Dataset} & \textbf{SR} & \textbf{RAG} & \textbf{SC} \\
\midrule
CounterBench    & $\uparrow$ 6.88 & $\uparrow$ 2.15 & $\uparrow$ 6.03\phantom{0} \\
GSM8K           & $\uparrow$ 0.82 & $\uparrow$ 0.10 & $\uparrow$ 0.72\phantom{0} \\
ReasonIF        & $\uparrow$ 8.59 & $\uparrow$ 0.03 & $\uparrow$ 11.89 \\
\midrule
\textbf{Average} & $\uparrow$ 5.43 & $\uparrow$ 0.76 & $\uparrow$ 6.21 \phantom{0} \\
\bottomrule
\end{tabular}
}
\caption{Accuracy gain from adding FLEx to each method. Values show average accuracy boost over the corresponding method alone, averaged across models.}
\label{tab:flex_method_delta}
\end{table}

\paragraph{Inference-time overhead.}
Because FLEx operates as a prompting method, it is important to quantify the additional cost introduced by prepending the selected summary $s^\star$. Across all datasets and model families, $s^\star$ adds an average of 162 tokens to the prompt (min--max: 73--301; Table~\ref{tab:inference_overhead_tokens}), which is modest relative to typical input lengths of several hundred tokens.

% Since inference latency in auto-regressive models is dominated by the forward pass over the full prompt, this additional cost scales linearly with prompt length and remains small in practice. 

\paragraph{Cross-model transfer of summaries.}
% To test whether FLEx distills model-agnostic counterfactual reasoning principles (rather than patching idiosyncratic behaviors of a single model), we perform a transfer experiment where the target model is fixed and only the learned summaries vary.
To test whether FLEx distills model-agnostic counterfactual reasoning principles rather than correcting idiosyncratic behaviors of a single model, we conduct a transfer experiment. We define the source model as the model used to collect explanations and construct the FLEx summary, and the target model as the model to which the summary is applied at inference-time. For each source model in the Gemma and Qwen families, we generate a FLEx summary $s^\star$ using CounterBench explanations collected from that source model. Each summary is then applied verbatim to a fixed target model (Qwen-7B) at inference-time. We evaluate transfer both in-domain on CounterBench and out-of-domain on CLadder \cite{jin2023cladder}, a separate counterfactual reasoning benchmark. 

\begin{table}[h]
\centering
\small
\setlength{\tabcolsep}{3pt}
\renewcommand{\arraystretch}{1.05}
\begin{tabular}{@{}lcc@{}}
\toprule
\textbf{Source Model} & \textbf{CounterBench} & \textbf{CLadder} \\
\midrule
Gemma-1B        & $\uparrow$ 0.9 & $\uparrow$ 1.44 \\
Gemma-4B        & $\downarrow$ 0.1 & $\uparrow$ 1.46 \\
Gemma-12B       & $\uparrow$ 4.1 & $\uparrow$ 1.78 \\
Gemma-27B       & $\uparrow$ 2.9 & $\uparrow$ 0.10 \\
Qwen-0.5B       & $\uparrow$ 5.4 & $\uparrow$ 1.77 \\
Qwen-1.5B       & $\uparrow$ 5.8 & $\uparrow$ 2.42 \\
Qwen-3B         & $\uparrow$ 4.6 & $\uparrow$ 2.29 \\
\textbf{Qwen-7B (self)}  & $\uparrow$ 8.1 & $\uparrow$ 1.45 \\
Qwen-14B        & $\uparrow$ 6.7 & $\uparrow$ 1.14 \\
Qwen-32B        & $\uparrow$ 2.1 & $\uparrow$ 0.19 \\
Qwen-72B        & $\uparrow$ 3.8 & $\uparrow$ 2.25 \\
\bottomrule
\end{tabular}
\caption{
Cross-model summary transfer to a fixed target model (Qwen-7B). Each summary $s^\star$ is distilled from CounterBench explanations for the \emph{source} model and applied verbatim to Qwen-7B at inference-time. We report accuracy change ($\Delta\text{Acc}$) on CounterBench (in-domain) and CLadder (out-of-domain).
}
\label{tab:cross_model_cladder}
\end{table}

Table~\ref{tab:cross_model_cladder} shows the results.
Despite being distilled from CounterBench errors produced by different source models, most summaries improve Qwen-7B's performance not only in-domain on CounterBench but also out-of-domain on CLadder. Transfer gains are generally strongest for summaries derived from models within the same family as the target model (Qwen) and from models of similar scale, though improvements are observed across a wide range of source architectures and sizes.

\section{Conclusion}

We introduced FLEx, a lightweight framework that improves reasoning performance by distilling a small number of verified, error-corrective explanations into a reusable prompt-level abstraction. Across diverse benchmarks and model scales, FLEx consistently improves over other training-free strategies, particularly in settings where model errors exhibit recurring patterns.  Our analysis shows that FLEx's gains arise from (i) selecting diverse errors based on model-internal similarity, (ii) using verified explanations rather than unvalidated explanations or corrected solutions, and (iii) distilling these explanations into concise, generalizable summaries.  More broadly, our results suggest that many reasoning failures reflect systematic misinterpretations rather than missing knowledge, and that targeted, human-interpretable explanations can effectively steer model behavior at inference-time.

\section*{Limitations}
While our method provides consistent improvements across models and datasets, several limitations remain. First, the approach relies on a relatively small set of selected error instances. Although the selection procedure aims to maximize diversity, a limited number of errors may not fully capture the breadth of model failures, especially for tasks with highly varied reasoning patterns. Second, the quality of the explanatory explanations depends on the annotators' ability to correctly diagnose and articulate the underlying error modes. Even with verification steps, human-written explanations may be partial, overly specific, or occasionally incorrect, which can constrain the generality of the resulting summaries. Finally, our evaluation focuses on single-turn reasoning tasks. Extending the method to multi-turn dialogue settings and to broader task families such as long-form generation, dialogue, or code synthesis remains for future work.

\section*{Ethical considerations}
This work aims to improve language modeling on various tasks and domains using language explanations provided by human annotators. Ensuring the fairness and robustness of these models is crucial to avoid biased or harmful outcomes, especially for underrepresented groups. Additionally, while our method enhances model performance, careful consideration is required before gathering explanations in order to avoid sensitive topics. Finally, the environmental impact of running inference on billion parameter language models is an important factor, and sustainable practices in AI research should be considered.

% \paragraph{Use of AI Assistants.}
% AI assistants were used to support code debugging and to improve the clarity and
% presentation of the manuscript. All experimental design, implementation, analysis,
% and conclusions were developed and verified by the authors.

% \section*{Acknowledgments}

% \clearpage
\bibliography{main}

@article{ouyang2022training,
  title={Training language models to follow instructions with human feedback},
  author={Ouyang, Long and Wu, Jeffrey and Jiang, Xu and Almeida, Diogo and Wainwright, Carroll and Mishkin, Pamela and Zhang, Chong and Agarwal, Sandhini and Slama, Katarina and Ray, Alex and others},
  journal={Advances in neural information processing systems},
  volume={35},
  pages={27730--27744},
  year={2022}
}

@article{devlin2018bert,
  title={Bert: Pre-training of deep bidirectional transformers for language understanding},
  author={Devlin, Jacob},
  journal={arXiv preprint arXiv:1810.04805},
  year={2018}
}

@article{radford2018improving,
  title={Improving language understanding by generative pre-training},
  author={Radford, Alec},
  journal={Advances in Neural Information Processing Systems},
  year={2018}
}

@article{chen2024learning,
  title={Learning from Natural Language Feedback},
  author={Chen, Angelica and Scheurer, J{\'e}r{\'e}my and Campos, Jon Ander and Korbak, Tomasz and Chan, Jun Shern and Bowman, Samuel R and Cho, Kyunghyun and Perez, Ethan},
  journal={Transactions on Machine Learning Research},
  year={2024}
}

@article{madaan2023self,
  title={Self-refine: Iterative refinement with self-feedback},
  author={Madaan, Aman and Tandon, Niket and Gupta, Prakhar and Hallinan, Skyler and Gao, Luyu and Wiegreffe, Sarah and Alon, Uri and Dziri, Nouha and Prabhumoye, Shrimai and Yang, Yiming and others},
  journal={Advances in Neural Information Processing Systems},
  volume={36},
  pages={46534--46594},
  year={2023}
}

@inproceedings{richardson2023syndicom,
  title={Syndicom: Improving Conversational Commonsense with Error-Injection and Natural Language Feedback},
  author={Richardson, Christopher and Sundar, Anirudh and Heck, Larry},
  booktitle={Proceedings of the 24th Annual Meeting of the Special Interest Group on Discourse and Dialogue},
  pages={297--308},
  year={2023}
}

@inproceedings{kwon2023efficient,
  title={Efficient Memory Management for Large Language Model Serving with PagedAttention},
  author={Woosuk Kwon and Zhuohan Li and Siyuan Zhuang and Ying Sheng and Lianmin Zheng and Cody Hao Yu and Joseph E. Gonzalez and Hao Zhang and Ion Stoica},
  booktitle={Proceedings of the ACM SIGOPS 29th Symposium on Operating Systems Principles},
  year={2023}
}

@article{vaswani2017attention,
  title={Attention is all you need},
  author={Vaswani, Ashish and Shazeer, Noam and Parmar, Niki and Uszkoreit, Jakob and Jones, Llion and Gomez, Aidan N and Kaiser, {\L}ukasz and Polosukhin, Illia},
  journal={Advances in neural information processing systems},
  volume={30},
  year={2017}
}

@article{brown2020language,
  title={Language models are few-shot learners},
  author={Brown, Tom and Mann, Benjamin and Ryder, Nick and Subbiah, Melanie and Kaplan, Jared D and Dhariwal, Prafulla and Neelakantan, Arvind and Shyam, Pranav and Sastry, Girish and Askell, Amanda and others},
  journal={Advances in neural information processing systems},
  volume={33},
  pages={1877--1901},
  year={2020}
}

@article{team2025gemma,
  title={Gemma 3 technical report},
  author={Team, Gemma and Kamath, Aishwarya and Ferret, Johan and Pathak, Shreya and Vieillard, Nino and Merhej, Ramona and Perrin, Sarah and Matejovicova, Tatiana and Ram{\'e}, Alexandre and Rivi{\`e}re, Morgane and others},
  journal={arXiv preprint arXiv:2503.19786},
  year={2025}
}

@article{yang2025qwen3,
  title={Qwen3 technical report},
  author={Yang, An and Li, Anfeng and Yang, Baosong and Zhang, Beichen and Hui, Binyuan and Zheng, Bo and Yu, Bowen and Gao, Chang and Huang, Chengen and Lv, Chenxu and others},
  journal={arXiv preprint arXiv:2505.09388},
  year={2025}
}

@inproceedings{rein2024gpqa,
  title={Gpqa: A graduate-level google-proof q\&a benchmark},
  author={Rein, David and Hou, Betty Li and Stickland, Asa Cooper and Petty, Jackson and Pang, Richard Yuanzhe and Dirani, Julien and Michael, Julian and Bowman, Samuel R},
  booktitle={First Conference on Language Modeling},
  year={2024}
}

@article{wei2022chain,
  title={Chain-of-thought prompting elicits reasoning in large language models},
  author={Wei, Jason and Wang, Xuezhi and Schuurmans, Dale and Bosma, Maarten and Xia, Fei and Chi, Ed and Le, Quoc V and Zhou, Denny and others},
  journal={Advances in neural information processing systems},
  volume={35},
  pages={24824--24837},
  year={2022}
}

@article{zelikman2022star,
  title={Star: Bootstrapping reasoning with reasoning},
  author={Zelikman, Eric and Wu, Yuhuai and Mu, Jesse and Goodman, Noah},
  journal={Advances in Neural Information Processing Systems},
  volume={35},
  pages={15476--15488},
  year={2022}
}

@article{shinn2023reflexion,
  title={Reflexion: Language agents with verbal reinforcement learning},
  author={Shinn, Noah and Cassano, Federico and Gopinath, Ashwin and Narasimhan, Karthik and Yao, Shunyu},
  journal={Advances in Neural Information Processing Systems},
  volume={36},
  pages={8634--8652},
  year={2023}
}

@article{bai2022constitutional,
  title={Constitutional ai: Harmlessness from ai feedback},
  author={Bai, Yuntao and Kadavath, Saurav and Kundu, Sandipan and Askell, Amanda and Kernion, Jackson and Jones, Andy and Chen, Anna and Goldie, Anna and Mirhoseini, Azalia and McKinnon, Cameron and others},
  journal={arXiv preprint arXiv:2212.08073},
  year={2022}
}

@article{huang2022large,
  title={Large language models can self-improve},
  author={Huang, Jiaxin and Gu, Shixiang Shane and Hou, Le and Wu, Yuexin and Wang, Xuezhi and Yu, Hongkun and Han, Jiawei},
  journal={arXiv preprint arXiv:2210.11610},
  year={2022}
}

@article{yao2023tree,
  title={Tree of thoughts: Deliberate problem solving with large language models},
  author={Yao, Shunyu and Yu, Dian and Zhao, Jeffrey and Shafran, Izhak and Griffiths, Tom and Cao, Yuan and Narasimhan, Karthik},
  journal={Advances in neural information processing systems},
  volume={36},
  pages={11809--11822},
  year={2023}
}

@article{wu2024discret,
  title={Discret: Synthesizing faithful explanations for treatment effect estimation},
  author={Wu, Yinjun and Keoliya, Mayank and Chen, Kan and Velingker, Neelay and Li, Ziyang and Getzen, Emily J and Long, Qi and Naik, Mayur and Parikh, Ravi B and Wong, Eric},
  journal={Proceedings of machine learning research},
  volume={235},
  pages={53597},
  year={2024}
}

@article{richardson2023integrating,
  title={Integrating summarization and retrieval for enhanced personalization via large language models},
  author={Richardson, Chris and Zhang, Yao and Gillespie, Kellen and Kar, Sudipta and Singh, Arshdeep and Raeesy, Zeynab and Khan, Omar Zia and Sethy, Abhinav},
  journal={arXiv preprint arXiv:2310.20081},
  year={2023}
}

@article{chen2025counterbench,
  title={Counterbench: A benchmark for counterfactuals reasoning in large language models},
  author={Chen, Yuefei and Singh, Vivek K and Ma, Jing and Tang, Ruxiang},
  journal={arXiv preprint arXiv:2502.11008},
  year={2025}
}

@article{cobbe2021training,
  title={Training verifiers to solve math word problems},
  author={Cobbe, Karl and Kosaraju, Vineet and Bavarian, Mohammad and Chen, Mark and Jun, Heewoo and Kaiser, Lukasz and Plappert, Matthias and Tworek, Jerry and Hilton, Jacob and Nakano, Reiichiro and others},
  journal={arXiv preprint arXiv:2110.14168},
  year={2021}
}

@article{kwon2025reasonif,
  title={ReasonIF: Large Reasoning Models Fail to Follow Instructions During Reasoning},
  author={Kwon, Yongchan and Zhu, Shang and Bianchi, Federico and Zhou, Kaitlyn and Zou, James},
  journal={arXiv preprint arXiv:2510.15211},
  year={2025}
}

@inproceedings{mcqueen1967some,
  title={Some methods of classification and analysis of multivariate observations},
  author={McQueen, James B},
  booktitle={Proc. of 5th Berkeley Symposium on Math. Stat. and Prob.},
  pages={281--297},
  year={1967}
}

@article{hurst2024gpt,
  title={Gpt-4o system card},
  author={Hurst, Aaron and Lerer, Adam and Goucher, Adam P and Perelman, Adam and Ramesh, Aditya and Clark, Aidan and Ostrow, AJ and Welihinda, Akila and Hayes, Alan and Radford, Alec and others},
  journal={arXiv preprint arXiv:2410.21276},
  year={2024}
}

@article{lanham2023measuring,
  title={Measuring faithfulness in chain-of-thought reasoning},
  author={Lanham, Tamera and Chen, Anna and Radhakrishnan, Ansh and Steiner, Benoit and Denison, Carson and Hernandez, Danny and Li, Dustin and Durmus, Esin and Hubinger, Evan and Kernion, Jackson and others},
  journal={arXiv preprint arXiv:2307.13702},
  year={2023}
}

@inproceedings{lampinen2022can,
  title={Can language models learn from explanations in context?},
  author={Lampinen, Andrew and Dasgupta, Ishita and Chan, Stephanie and Mathewson, Kory and Tessler, Mh and Creswell, Antonia and McClelland, James and Wang, Jane and Hill, Felix},
  booktitle={Findings of the Association for Computational Linguistics: EMNLP 2022},
  pages={537--563},
  year={2022}
}

@article{turpin2023language,
  title={Language models don't always say what they think: Unfaithful explanations in chain-of-thought prompting},
  author={Turpin, Miles and Michael, Julian and Perez, Ethan and Bowman, Samuel},
  journal={Advances in Neural Information Processing Systems},
  volume={36},
  pages={74952--74965},
  year={2023}
}

@article{wang2022self,
  title={Self-consistency improves chain of thought reasoning in language models},
  author={Wang, Xuezhi and Wei, Jason and Schuurmans, Dale and Le, Quoc and Chi, Ed and Narang, Sharan and Chowdhery, Aakanksha and Zhou, Denny},
  journal={arXiv preprint arXiv:2203.11171},
  year={2022}
}

@article{jin2023cladder,
  title={Cladder: Assessing causal reasoning in language models},
  author={Jin, Zhijing and Chen, Yuen and Leeb, Felix and Gresele, Luigi and Kamal, Ojasv and Lyu, Zhiheng and Blin, Kevin and Gonzalez Adauto, Fernando and Kleiman-Weiner, Max and Sachan, Mrinmaya and others},
  journal={Advances in Neural Information Processing Systems},
  volume={36},
  pages={31038--31065},
  year={2023}
}

@article{ouyang2025reasoningbank,
  title={Reasoningbank: Scaling agent self-evolving with reasoning memory},
  author={Ouyang, Siru and Yan, Jun and Hsu, I and Chen, Yanfei and Jiang, Ke and Wang, Zifeng and Han, Rujun and Le, Long T and Daruki, Samira and Tang, Xiangru and others},
  journal={arXiv preprint arXiv:2509.25140},
  year={2025}
}

@article{clark2018think,
  title={Think you have solved question answering? try arc, the ai2 reasoning challenge},
  author={Clark, Peter and Cowhey, Isaac and Etzioni, Oren and Khot, Tushar and Sabharwal, Ashish and Schoenick, Carissa and Tafjord, Oyvind},
  journal={arXiv preprint arXiv:1803.05457},
  year={2018}
}

@article{hendrycks2021measuring,
  title={Measuring mathematical problem solving with the math dataset},
  author={Hendrycks, Dan and Burns, Collin and Kadavath, Saurav and Arora, Akul and Basart, Steven and Tang, Eric and Song, Dawn and Steinhardt, Jacob},
  journal={arXiv preprint arXiv:2103.03874},
  year={2021}
}

@inproceedings{satopaa2011finding,
  title={Finding a" kneedle" in a haystack: Detecting knee points in system behavior},
  author={Satopaa, Ville and Albrecht, Jeannie and Irwin, David and Raghavan, Barath},
  booktitle={2011 31st international conference on distributed computing systems workshops},
  pages={166--171},
  year={2011},
  organization={IEEE}
}

@inproceedings{wolf2020transformers,
  title={Transformers: State-of-the-art natural language processing},
  author={Wolf, Thomas and Debut, Lysandre and Sanh, Victor and Chaumond, Julien and Delangue, Clement and Moi, Anthony and Cistac, Pierric and Rault, Tim and Louf, Remi and Funtowicz, Morgan and others},
  booktitle={Proceedings of the 2020 conference on empirical methods in natural language processing: system demonstrations},
  pages={38--45},
  year={2020}
}

@inproceedings{besta2024graph,
  title={Graph of thoughts: Solving elaborate problems with large language models},
  author={Besta, Maciej and Blach, Nils and Kubicek, Ales and Gerstenberger, Robert and Podstawski, Michal and Gianinazzi, Lukas and Gajda, Joanna and Lehmann, Tomasz and Niewiadomski, Hubert and Nyczyk, Piotr and others},
  booktitle={Proceedings of the AAAI conference on artificial intelligence},
  volume={38},
  number={16},
  pages={17682--17690},
  year={2024}
}

@article{turner2024activation,
  title={Activation addition: Steering language models without optimization},
  author={Turner, Alexander Matt and Thiergart, Lisa and Leech, Gavin and Udell, David and Mini, Ulisse and MacDiarmid, Monte},
  year={2024}
}

@article{lewis2020retrieval,
  title={Retrieval-augmented generation for knowledge-intensive nlp tasks},
  author={Lewis, Patrick and Perez, Ethan and Piktus, Aleksandra and Petroni, Fabio and Karpukhin, Vladimir and Goyal, Naman and K{\"u}ttler, Heinrich and Lewis, Mike and Yih, Wen-tau and Rockt{\"a}schel, Tim and others},
  journal={Advances in neural information processing systems},
  volume={33},
  pages={9459--9474},
  year={2020}
}

@article{zhou2022least,
  title={Least-to-most prompting enables complex reasoning in large language models},
  author={Zhou, Denny and Sch{\"a}rli, Nathanael and Hou, Le and Wei, Jason and Scales, Nathan and Wang, Xuezhi and Schuurmans, Dale and Cui, Claire and Bousquet, Olivier and Le, Quoc and others},
  journal={arXiv preprint arXiv:2205.10625},
  year={2022}
}

\clearpage
\appendix \label{sec:appendix}
\section{Prompts}
\label{app:prompts-eval}

This section documents the exact prompting, answer extraction, and
evaluation procedures used for the experiments. Each input consists of a system prompt and a user prompt. When FLEx is enabled, the selected summary $s^\star$ is appended verbatim to the system prompt. No other changes are made to decoding or task text. Below, \texttt{\{TASK\}} denotes the dataset-provided input, and \texttt{\{SUMMARY\}} denotes $s^\star$ (empty for non-FLEx baselines).

\paragraph{Base prompts.}
\label{app:base_prompts}

\subparagraph{CounterBench}\mbox{}\\[.1em]
\par\vspace{-0.5\baselineskip}

\begin{tcolorbox}
\textbf{System:} You are a helpful assistant. Show your reasoning step by step.
At the end, output <answer>yes</answer> or <answer>no</answer>.
\{SUMMARY\}
\end{tcolorbox}

\vspace{6pt}
\begin{tcolorbox}
\textbf{User:} \{TASK\}
\end{tcolorbox}

\subparagraph{GSM8K}\mbox{}\\[.1em]
\par\vspace{-0.5\baselineskip}

\begin{tcolorbox}
\textbf{System:} You are a helpful assistant. Show your reasoning step by step. At the end, output the final answer in <answer> tags. \{SUMMARY\}
\end{tcolorbox}

\vspace{6pt}
\begin{tcolorbox}
\textbf{User:} \{TASK\}
\end{tcolorbox}

\subparagraph{ReasonIF}\mbox{}\\[.1em]
\par\vspace{-0.5\baselineskip}

\begin{tcolorbox}
\textbf{System:} You are a helpful assistant. Show your reasoning step by step. At the end, output the final answer in <answer> tags. \{SUMMARY\}
\end{tcolorbox}

\vspace{6pt}
\begin{tcolorbox}
\textbf{User:} \{TASK\}
\end{tcolorbox}

\vspace{8pt}
For all datasets, we extract the content of the final \texttt{<answer>...</answer>} tag (case-insensitive). If no such tag is present, the prediction is marked incorrect. Extracted answers are normalized by stripping capitalization, punctuation, and symbols.

\paragraph{Self-Refine.}
\label{app:self_refine}

Self-refine is a two-stage procedure applied to a baseline model output $y_0$.

\begin{enumerate}
    \item \textbf{Changes:} the model reviews $(\texttt{\{TASK\}}, y_0)$ and outputs either \texttt{None} or a bullet list of required edits.
    \item \textbf{Revise:} if Changes $\neq$ \texttt{None}, the model edits $y_0$ using the proposed changes to produce $y_1$.
\end{enumerate}

If Changes = \texttt{None}, we keep $y_0$. If a revision is attempted but the revised output violates the required format, we fall back to $y_0$. Below, \texttt{\{TASK\}} denotes the dataset-provided input, and \texttt{\{DRAFT\}} denotes the model's initial response $y_0$. The placeholder \texttt{\{FORMAT\}} expands to the dataset-specific output constraint.

\paragraph{Changes.}\mbox{}\\[.1em]
\par\vspace{-0.5\baselineskip}

\begin{tcolorbox}
\textbf{System:} You are a careful self-reviewer.
Output EXACTLY 'NONE' or a bullet list.
Never include <answer> tags in this step.
Do not rewrite the full answer.
\end{tcolorbox}

\vspace{6pt}
\begin{tcolorbox}
\textbf{User:} You are reviewing your own draft answer.
Your task: identify only what must change to make the draft correct and compliant.
If nothing needs to change, output exactly:
none. Otherwise output a bullet list where each line starts with '- '.
Do NOT rewrite the answer.

REQUIRED OUTPUT FORMAT FOR THE FINAL ANSWER:
\{FORMAT\}

USER PROMPT:
\{TASK\}

DRAFT RESPONSE:
\{DRAFT\}

CHANGES:
\end{tcolorbox}

\paragraph{Revise.}\mbox{}\\[.1em]
\par\vspace{-0.5\baselineskip}

\begin{tcolorbox}
\textbf{System:} You are revising your own answer.
Output ONLY the revised response.
Follow the required output format exactly.
\end{tcolorbox}

\vspace{6pt}
\begin{tcolorbox}
\textbf{User:} Revise the draft by applying the REQUIRED CHANGES.
Output ONLY the revised response. No commentary.
You MUST follow the REQUIRED OUTPUT FORMAT exactly.

REQUIRED OUTPUT FORMAT:
\{FORMAT\}

USER PROMPT:
\{TASK\}

ORIGINAL DRAFT:
\{DRAFT\}

REQUIRED CHANGES:
\{CHANGES\}

REVISED RESPONSE:
\end{tcolorbox}
\vspace{8pt}

\paragraph{Retrieval-augmented generation (RAG).}
\label{app:rag}

For the RAG baseline, we build a retrieval index over the model's correct training examples for each dataset and model. Each indexed item consists of the original task input paired with the model's full solution. We index only the task input text, excluding answers and reasoning. Inputs and queries are encoded using a frozen SentenceTransformer model (\texttt{all-MiniLM-L6-v2}) with $\ell_2$-normalized embeddings. Similarity search is performed using FAISS inner-product search (cosine similarity). At test-time, we retrieve the single nearest neighbor ($k=1$) from the training index and prepend its prompt and solution verbatim to the user prompt. Indices are constructed exclusively from training data; no test examples are indexed. System prompts for each dataset are the same as in the base prompts above (Appendix~\ref{app:base_prompts}). User prompts are provided below.

\vspace{6pt}
\begin{tcolorbox}
\textbf{User:}
\\ Question:
\{RETRIEVED\_PROMPT\}

Solution:
\{RETRIEVED\_RESPONSE\}

Question:
\{TASK\}

Solution:
\end{tcolorbox}

\paragraph{Self-Consistency.}

When self-consistency is enabled, we sample $n=5$ independent responses with temperature $0.7$ and select the majority answer after parsing. Ties fall back to the first sample. All prompts are the same as above (Appendix~\ref{app:base_prompts}).

\section{Compute}
\begin{table}[ht]
    \centering
    \begin{tabularx}{\columnwidth}{Xc}
        \toprule
        \textbf{Parameter} & \textbf{Value} \\
        \midrule
        temperature & 0.0 \\
        num\_beams & 1 \\
        top\_k & disabled ($-1$) \\
        top\_p & 1.0 \\
        repetition\_penalty & 1.0 \\
        max\_new\_tokens & 8192 \\
        stop\_sequences & none \\
        \bottomrule
    \end{tabularx}
    \caption{Hyperparameters used for decoding.}
    \label{tab:hyperparameters}
\end{table}
We use vLLM \cite{kwon2023efficient} for inference. Parameters are shown in Table \ref{tab:hyperparameters}. All inference was run on Nvidia A40 GPUs with 48GB GDDR6 memory. We use vLLM tensor parallelism across 2, 4, or 8 GPUs depending on model size. Runtime ranged from a few minutes to about one hour per (model, dataset) evaluation, depending on model size and dataset.

% \begin{table}[h]
%     \centering
%     \begin{tabularx}{\columnwidth}{Xl}
%         \toprule
%         \textbf{Model} & \textbf{Special Tokens} \\ 
%         \midrule

%         \multirow{4}{*}{Gemma} 
%         & \texttt{<bos>} \\
%         & \texttt{<start\_of\_turn>} \\
%         & \texttt{<end\_of\_turn>} \\
%         & \texttt{<eos>} \\
%         \midrule

%         \multirow{6}{*}{Qwen} 
%         & \texttt{<bos>} \\
%         & \texttt{<pad>} \\
%         & \texttt{<unk>} \\
%         & \texttt{<|im\_start|>} \\
%         & \texttt{<|im\_end|>} \\
%         & \texttt{<eos>} \\
%         \bottomrule

%     \end{tabularx}
%     \caption{Special tokens used for Gemma and Qwen models.}
%     \label{tab:special_tokens}
% \end{table}

\subsection{Models}
\label{app:models}

We evaluate instruction-tuned models from the Gemma-3 and Qwen-2.5 families, using the following Hugging Face checkpoints without modification \cite{wolf2020transformers}.

% \begin{table}[H]
% \centering
% \resizebox{\columnwidth}{!}{
% \begin{tabular}{l l}
% \toprule
% \textbf{Family} & \textbf{Hugging Face Model ID} \\
% \midrule
% Gemma-3 & \texttt{google/gemma-3-1b-it} \\
%         & \texttt{google/gemma-3-4b-it} \\
%         & \texttt{google/gemma-3-12b-it} \\
%         & \texttt{google/gemma-3-27b-it} \\
% \midrule
% Qwen-2.5 & \texttt{Qwen/Qwen2.5-0.5B-Instruct} \\
%          & \texttt{Qwen/Qwen2.5-1.5B-Instruct} \\
%          & \texttt{Qwen/Qwen2.5-3B-Instruct} \\
%          & \texttt{Qwen/Qwen2.5-7B-Instruct} \\
%          & \texttt{Qwen/Qwen2.5-14B-Instruct} \\
%          & \texttt{Qwen/Qwen2.5-32B-Instruct} \\
%          & \texttt{Qwen/Qwen2.5-72B-Instruct} \\
% \bottomrule
% \end{tabular}
% }
% \caption{Exact Hugging Face model checkpoints used in our experiments.}
% \end{table}

\begin{table}[H]
\centering
\scriptsize
\setlength{\tabcolsep}{4pt}
\renewcommand{\arraystretch}{1.05}
\begin{tabularx}{\columnwidth}{l l X}
\toprule
\textbf{Family} & \textbf{Hugging Face Model ID} & \textbf{Special tokens} \\
\midrule
Gemma-3
& \texttt{google/gemma-3-1b-it}
& \multirow{4}{=}{\texttt{<bos>}, \texttt{<start\_of\_turn>}, \texttt{<end\_of\_turn>}, \texttt{<eos>}} \\
& \texttt{google/gemma-3-4b-it} & \\
& \texttt{google/gemma-3-12b-it} & \\
& \texttt{google/gemma-3-27b-it} & \\
\midrule
Qwen-2.5
& \texttt{Qwen/Qwen2.5-0.5B-Instruct}
& \multirow{7}{=}{\texttt{<bos>}, \texttt{<pad>}, \texttt{<unk>}, \texttt{<|im\_start|>}, \texttt{<|im\_end|>}, \texttt{<eos>}} \\
& \texttt{Qwen/Qwen2.5-1.5B-Instruct} & \\
& \texttt{Qwen/Qwen2.5-3B-Instruct} & \\
& \texttt{Qwen/Qwen2.5-7B-Instruct} & \\
& \texttt{Qwen/Qwen2.5-14B-Instruct} & \\
& \texttt{Qwen/Qwen2.5-32B-Instruct} & \\
& \texttt{Qwen/Qwen2.5-72B-Instruct} & \\
\bottomrule
\end{tabularx}
\caption{Instruction-tuned Gemma-3 and Qwen-2.5 models evaluated, with special tokens used for prompt construction.}
\label{tab:models_and_special_tokens}
\end{table}

\paragraph{Summary generation model (OpenAI).}
We used the OpenAI Chat Completions API with the model alias \texttt{gpt-4o-mini} with temperature \texttt{T=1.0}.

\subsection{Datasets}
\label{sec:datasets}

We conduct experiments on four reasoning benchmarks: CounterBench, GSM8K, ReasonIF, and CLadder. For CounterBench, we define fixed train and test splits using \texttt{data\_balanced\_backdoor\_V2.json} (train) and \texttt{data\_balanced\_alpha\_V1.json} (test). GSM8K uses its standard train/test split. For ReasonIF, we synthetically generate a small training split for explanation selection, as described in Appendix~\ref{sec:reasonif}. For CLadder, we perform evaluation exclusively on the test split and do not use any training data.

\begin{table}[H]
\centering
\scriptsize
\setlength{\tabcolsep}{4pt}
\renewcommand{\arraystretch}{1.05}
\begin{tabularx}{\columnwidth}{l >{\raggedright\arraybackslash}X r r}
\toprule
\textbf{Dataset} & \textbf{HF ID} & \textbf{Train} & \textbf{Test} \\
\midrule
CounterBench & \texttt{CounterBench/CounterBench} & 200 & 1000 \\
GSM8K        & \texttt{openai/gsm8k}              & 7473 & 1319 \\
ReasonIF     & \texttt{ykwon-hf/reasonIF}         & 100 & 300 \\
CLadder      & \texttt{causal-nlp/CLadder}        & -- & 10112 \\
\bottomrule
\end{tabularx}
\caption{Datasets used in our experiments and their split sizes.}
\label{tab:datasets}
\end{table}

\section{Clustering Details}
\label{app:clustering_details}
To select a small but diverse set of model errors for annotation, we cluster incorrect examples produced by each model on each dataset. For each error $(x, r)$, we embed the concatenated sequence $[x; r]$ using the last-token hidden state from the final transformer layer of the same frozen model that generated the error. This ensures that clustering reflects the model's own internal error representations.

We apply $k$-means clustering (with $k$-means++ initialization) to these embeddings in order to group errors into
distinct failure modes. Let $\{\mathbf{e}_i\}_{i=1}^n \subset \mathbb{R}^d$ denote the embeddings of $n$ erroneous examples. For a given number of clusters $k$, $k$-means partitions the embeddings into clusters
$\{\mathcal{C}_j\}_{j=1}^k$ with centroids $\{\boldsymbol{\mu}_j\}_{j=1}^k$ by minimizing the within-cluster sum of squares (inertia):
\[
\mathcal{I}(k) = \sum_{j=1}^{k} \sum_{\mathbf{e}_i \in \mathcal{C}_j}
\left\lVert \mathbf{e}_i - \boldsymbol{\mu}_j \right\rVert_2^2 .
\]

To determine the appropriate number of clusters, we use the elbow method based on inertia. We sweep $k \in [2, \min(20, n)]$ and compute $\mathcal{I}(k)$ for each value. Let $\{(k, \mathcal{I}(k))\}$ denote the resulting curve. We select $k^\star$ using the maximum distance to chord criterion \cite{satopaa2011finding}, which chooses the point on the inertia curve with the largest perpendicular distance to the line connecting its endpoints, corresponding to the knee of the curve.  Intuitively, this corresponds to the point beyond which additional clusters yield diminishing reductions in within-cluster variance.

\begin{figure}
    \centering
    \includegraphics[width=1\linewidth]{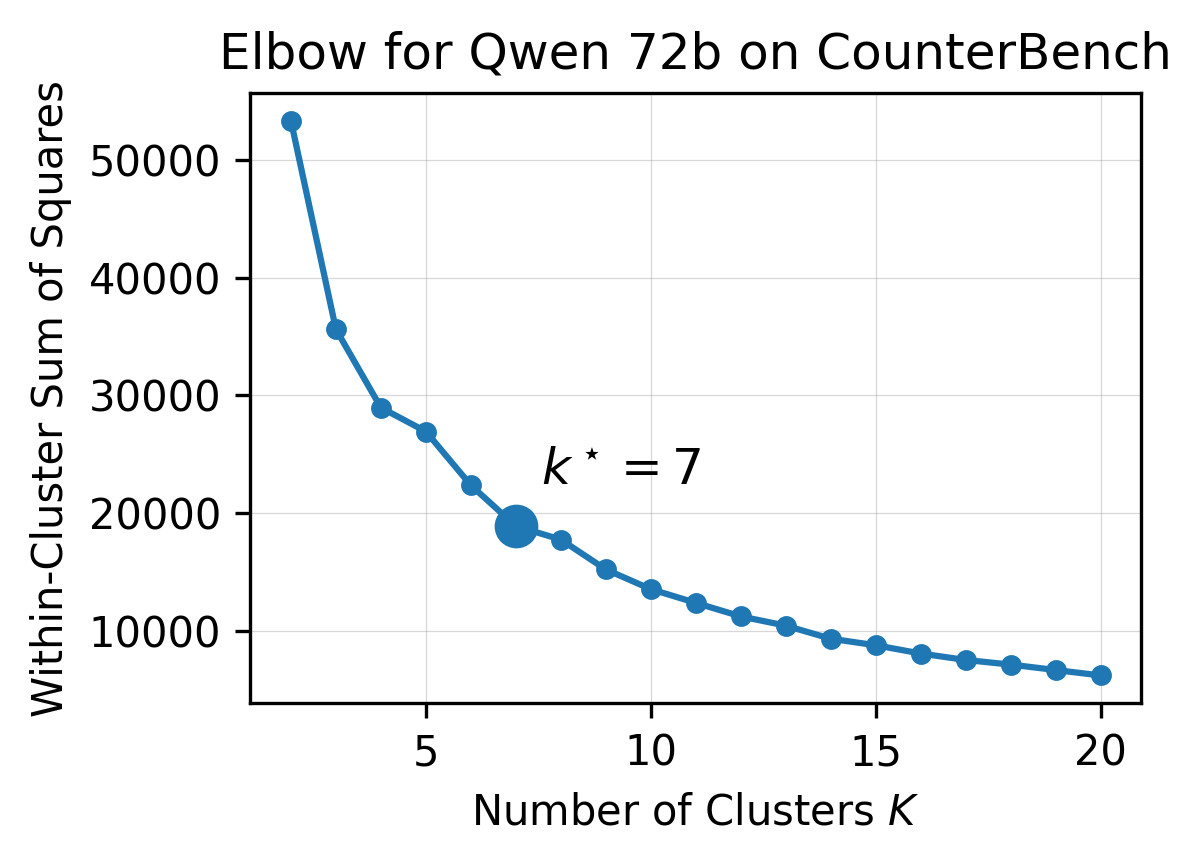}
    \caption{Elbow Plot for Qwen 72b on Counterbench}
    \label{fig:placeholder}
\end{figure}

Figure~\ref{fig:placeholder} shows an example elbow plot for Qwen~72B on CounterBench. In this case, the automatically selected $k^\star$ coincides with the visually apparent ``elbow'' of the curve, where further increases in $k$ yield only marginal reductions in within-cluster variance, consistent with standard heuristic inspection of elbow plots.

Once $k^\star$ is selected, we rerun $k$-means with $k^\star$ clusters and select one representative error per cluster-the instance closest to the cluster centroid-as the primary candidate for annotation. Additional nearest neighbors are retained as backups if the primary example cannot be successfully verified. Table~\ref{tab:kstar_full} reports the selected number of clusters $k^\star$ for each model--dataset pair.

\begin{table}[H]
\centering
\resizebox{\columnwidth}{!}{
\begin{tabular}{lcccc}
\toprule
\textbf{Model} & \textbf{CounterBench} & \textbf{GSM8K}  & \textbf{ReasonIF} \\
\midrule
Gemma 1B      & 10 & 7 & 7 \\
Gemma 4B      & 7 & 8 & 5 \\
Gemma 12B     & 8 & 11 & 7 \\
Gemma 27B     & 7 & 9 & 8 \\
Qwen 0.5B     & 4 & 6 & 11 \\
Qwen 1.5B     & 8 & 9 & 6 \\
Qwen 3B       & 4 & 5 & 6 \\
Qwen 7B       & 8 & 8 & 7 \\
Qwen 14B      & 5 & 7 & 6 \\
Qwen 32B      & 6 & 8 & 5 \\
Qwen 72B      & 7 & 7 & 6 \\
\bottomrule
\end{tabular}
}
\caption{Selected number of clusters $k^*$ for each model-dataset pair, computed using the elbow method.}
\label{tab:kstar_full}
\end{table}

We observe substantial variation in $k^\star$ across models and tasks: smaller models and more heterogeneous datasets tend to produce a broader distribution of error types, resulting in larger values of $k$. These clusters form the basis for selecting the small set of verified explanations used by FLEx.

\section{ReasonIF Training Set Construction}
\label{sec:reasonif}

We construct a synthetic ReasonIF training set of 100 examples to match the test
distribution.
\begin{itemize}
    \item Data is drawn from GSM8K, AMC, AIME, GPQA, and ARC \cite{cobbe2021training, hendrycks2021measuring, rein2024gpqa, clark2018think}.
    \item Exact question overlap with test set is removed. 
    \item Constraint types and argument distributions are matched to the test split.
    \item Prompts are formatted identically to the test-set.

\end{itemize}
This ensures that explanation annotation and summarization reflect the same constraint structure encountered at evaluation time.

\section{Explanation Examples and Summaries}

We present end-to-end examples for CounterBench, GSM8K, and ReasonIF (Tables \ref{tab:counterbench}, \ref{tab:gsm8k}, and \ref{tab:reasonif}) that illustrate the full pipeline from an initial incorrect model response, through human explanations, to the corrected response after applying the explanations. We additionally report the summary prompts used to generate candidate summaries (Table \ref{tab:summary_prompts}), along with representative example summaries produced by each prompt (Table \ref{tab:summary_examples}).

\section{Complete Ablation Results}
\label{sec:ablations}
We provide here the full set of results discussed in the analysis section of the main text, including both ablation studies and additional experimental variants.

\onecolumn

\begin{table}[!htbp]
\centering
\small
\setlength{\tabcolsep}{4pt}
\renewcommand{\arraystretch}{0.95}
\begin{tabularx}{\textwidth}{X}
\toprule
\textbf{CounterBench End-to-End Example} \\
\midrule
\textbf{Prompt:} We know that Blorn causes Fizo, Fizo or Blorn causes Plim, Plim causes Quaz, Quaz causes Skul, and Skul causes Jext. Blorn~Bern(0.4). We observed Plim. Would Jext occur if not Fizo instead of Fizo? \\[1.5pt]

\textbf{Incorrect model response:}
To determine if Jext would occur if not Fizo instead of Fizo, we need to follow the causal chain and understand the implications of the absence of Fizo.
% \begin{enumerate}\itemsep0pt
% \item \textbf{Causal chain analysis:} Blorn causes Fizo. Fizo or Blorn causes Plim. Plim causes Quaz. Quaz causes Skul. Skul causes Jext.
% \item \textbf{Given information:} Blorn~Bern(0.4) means Blorn has a 40\% chance of occurring. We observed Plim.
% \item \textbf{Implications of observing Plim:} Plim can be caused by either Fizo or Blorn.
% \item \textbf{Scenario analysis:} 
% \bad{If Fizo did not occur, Blorn must have occurred. Since Blorn causes Fizo, Fizo would still occur.}
% \item \textbf{Effect on Jext:} The causal chain remains intact, so Jext would still occur.
% \end{enumerate}

Blorn causes Fizo. Fizo or Blorn causes Plim. Plim causes Quaz. Quaz causes Skul. Skul causes Jext. Blorn~Bern(0.4) means Blorn has a 40\% chance of occurring. We observed Plim. Plim can be caused by either Fizo or Blorn. \bad{If Fizo did not occur, Blorn must have occurred. Since Blorn causes Fizo, Fizo would still occur.} The causal chain remains intact, so Jext would still occur. Therefore, the answer is <answer>no</answer>. \\[1.5pt]

\textbf{Human-Written Explanation:}
Your error is that you assumed Plim requires Fizo, or that Blorn $\rightarrow$ Fizo makes Fizo unavoidable. The counterfactual instead asks whether Jext would still occur if Fizo were explicitly prevented. \\[1.5pt]

\textbf{Corrected Response:}
\good{Under the counterfactual where Fizo is prevented, Plim can still be caused by Blorn, allowing the causal chain to continue to Jext.}
Therefore, the correct answer is <answer>yes</answer>. \\
\bottomrule
\end{tabularx}
\caption{End-to-end example for CounterBench.}
\label{tab:counterbench}
\end{table}

\begin{table}[!htbp]
\centering
\small
\setlength{\tabcolsep}{4pt}
\renewcommand{\arraystretch}{0.95}
\begin{tabularx}{\textwidth}{X}
\toprule
\textbf{GSM8K End-to-End Example} \\
\midrule
\textbf{Prompt:} There are 88 dogs in a park. 12 of the dogs are running. Half of them are playing with toys. A fourth of them are barking. How many dogs are not doing anything? \\[1.5pt]

\textbf{Incorrect model response:}
\begin{enumerate}[nosep,topsep=0pt,partopsep=0pt,parsep=0pt,leftmargin=*]
\item Dogs running: 12
\item Dogs playing with toys: \bad{$12 / 2 = 6$}
\item Dogs barking: \bad{$12 / 4 = 3$}
\item Total dogs doing something: $12 + 6 + 3 = 21$
\item Dogs doing nothing: $88 - 21 = 67$
\end{enumerate}
<answer>67</answer> \\[1.5pt]

\textbf{Human-Written Explanation:}
You misread the problem. ``Half of them'' and ``a fourth of them'' refer to half and a fourth of all 88 dogs, not of the 12 running dogs. \\[1.5pt]

\textbf{Corrected Response:}
\begin{enumerate}[nosep,topsep=0pt,partopsep=0pt,parsep=0pt,leftmargin=*]
\item Dogs running: 12
\item Dogs playing with toys: \good{$88 / 2 = 44$}
\item Dogs barking: \good{$88 / 4 = 22$}
\item Total dogs doing something: $12 + 44 + 22 = 78$
\item Dogs doing nothing: $88 - 78 = 10$
\end{enumerate}
<answer>10</answer> \\
\bottomrule
\end{tabularx}
\caption{End-to-end example for GSM8K.}
\label{tab:gsm8k}
\end{table}

\begin{table}[!htbp]
\centering
\small
\setlength{\tabcolsep}{4pt}
\renewcommand{\arraystretch}{0.95}
\begin{tabularx}{\textwidth}{X}
\toprule
\textbf{ReasonIF End-to-End Example} \\
\midrule
\textbf{Prompt:} When reasoning, your response should be in English and in all capital letters. Here is the question:
Triangle $ABC$ has side lengths in arithmetic progression, and the smallest side has length $6.$ If the triangle has an angle of $120^{\circ},$ find the area of $ABC$. The final answer can be simplified in the form $m \sqrt{n}$, where $m$ and $n$ are positive integers and $n$ has no square factor. What is $m+n$? \\[1.5pt]

\textbf{Incorrect model response:}
LET THE SIDE LENGTHS BE $6, 6+\badmath{d}, 6+2\badmath{d}$ FOR SOME $\badmath{d}>0$.  
SINCE THE SIDES ARE IN ARITHMETIC PROGRESSION, WE HAVE $6 < 6+\badmath{d} < 6+2\badmath{d}$.  
IF THE ANGLE OPPOSITE THE SIDE OF LENGTH $6+2\badmath{d}$ IS $120^{\circ}$, THEN  
\[
6^2 + (6+\badmath{d})^2 - 2(6)(6+\badmath{d}) \badmath{\cos}(120^{\circ}) = (6+2\badmath{d})^2.
\]
SOLVING THIS EQUATION YIELDS $\badmath{d}=4$.  
THE SIDE LENGTHS ARE $6, 10, 14$.  
THE AREA IS $\frac{1}{2}(6)(10)\badmath{\sin}(120^{\circ}) = 15\sqrt{3}$.  
THUS, $\badmath{m}+\badmath{n} = 18$.  
<answer>18</answer> \\[1.5pt]

\textbf{Human-Written Explanation:}
Your response violates the formatting constraint: mathematical variables and expressions contain lowercase letters (e.g., $d>0$). All text, including mathematical notation, must be capitalized. \\[1.5pt]

\textbf{Corrected Response:}
LET THE SIDE LENGTHS BE 6, $6+\goodmath{D}, 6+2D$ FOR SOME \goodmath{D}>0.  
IF THE ANGLE OPPOSITE THE SIDE OF LENGTH $6+2\goodmath{D}$ IS 120 DEGREES, THEN  
\[
6^2 + (6+\goodmath{D})^2 - 2(6)(6+\goodmath{D})\goodmath{\text{COS}}(120) = (6+2\goodmath{D})^2.
\]
SOLVING GIVES \goodmath{D}=4.  
THE SIDE LENGTHS ARE 6, 10, AND 14.  
THE AREA IS $(1/2)(6)(10)\goodmath{\text{SIN}}(120) = 15\text{ SQRT}(3)$.  
THUS, M=15, N=3, AND \goodmath{M}+\goodmath{N} = 18.  
<answer>18</answer> \\
\bottomrule
\end{tabularx}
\caption{End-to-end example for ReasonIF.}
\label{tab:reasonif}
\end{table}

% \FloatBarrier

\begin{table}[H]
\centering
\small
\setlength{\tabcolsep}{6pt}
\renewcommand{\arraystretch}{1.1}
\begin{tabularx}{\textwidth}{l X c c}
\toprule
\textbf{Name} & \textbf{Summary Prompt} & \textbf{Source} & \textbf{Best (\#)} \\
\midrule
Bullet Rules &
From the following feedback, extract the most important ideas a model should remember to avoid similar mistakes again. Present them as clear bullet points.
& Author & 32\% \\
\midrule

LLM Commands &
Given prompt-response-feedback triples, infer general behavioral rules that raised scores. Summarize them as direct, command-style instructions. Use imperative tone, exaggerate compliance, forbid reasoning or commentary, and keep output JSON-safe and under 120 tokens.    
 & LLM &
31\% \\
\midrule

Crisp Lessons & 
Summarize the feedback into a few crisp and actionable lessons that could help improve similar responses in the future. Focus on clarity and utility.    
 & Author &
24\% \\
\midrule

Single Directive & 
Summarize feedback patterns into a single behavioral directive for LLM inference. Focus on strict obedience, clear polarity, and exact format replication. Output short, direct, rule-like instructions only.   
 & Author &
10\% \\
\midrule

LLM Paragraph & 
You are analyzing multiple (prompt, response, feedback) triplets. Produce a single summary that teaches another model how to perform better on future tasks of the same kind. Identify repeated behavioral corrections, extract underlying rules that improve evaluation scores, and rewrite them as short, imperative, command-style instructions. Use exaggerated compliance for limits, avoid meta-phrasing, keep the output deterministic and JSON-safe, and compress it into one paragraph under 120 tokens. Optionally include a compact version under 40 tokens.
 & LLM &
3\% \\

\bottomrule
\end{tabularx}
\caption{Summary prompt variants used to generate candidate summaries. The final column reports how often each prompt produced the top-ranked summary according to the $\Delta$-embedding similarity metric.}
\label{tab:summary_prompts}
\end{table}

\begin{table}[H]
\centering
\small
\setlength{\tabcolsep}{6pt}
\renewcommand{\arraystretch}{1.15}
\begin{tabularx}{\textwidth}{l p{0.7\textwidth}}
\toprule
\textbf{Prompt} & \textbf{Example Generated Summary} \\
\midrule

Bullet Rules &
\begin{itemize}[leftmargin=*, labelsep=0.5em, nosep]
  \item Avoid contradictory answers: Ensure the conclusion matches the final answer.
  \item Understand OR logic in causation: If a causal statement uses ``or'', either condition can sustain causation.
  \item Properly trace counterfactuals: When evaluating ``if not X,'' set X false and follow implications.
  \item Utilize observations: Observed variables constrain the counterfactual world unless intervened on.
  \item Avoid irrelevant probabilistic reasoning: Do not introduce probabilities in deterministic settings.
\end{itemize}
\\
\midrule

LLM Commands &
\{
  "rules": [
    "If a node is observed, hold it true in counterfactuals.",
    "Trace causal chains accurately without assuming absent causes can still yield observed outcomes.",
    "Always implement counterfactuals by altering upstream nodes only.",
    "Recompute all downstream effects based on altered conditions.",
    "If a path to the target node exists, return yes; if it's broken, return no."
  ]
\}
\\
\midrule

Crisp Lessons &
Actionable Lessons:
\begin{enumerate}[leftmargin=*, labelsep=0.5em, nosep]
  \item Counterfactual framework: Observed nodes remain true unless intervened on.
  \item Clarify causal relationships: Treat ``A or B'' as independently sufficient.
  \item Conditional logic: Allow alternative paths to preserve downstream effects.
  \item Precision in conclusions: Avoid ``cannot occur'' when observations support occurrence.
  \item Focus on the query: Apply the intervention and answer directly.
\end{enumerate}
\\
\midrule

Single directive & \begin{enumerate}[leftmargin=*, labelsep=0.5em, nosep]
  \item Treat causes as sufficient but not necessary; avoid false equivalences.
  \item Use observations to infer the factual world before intervention.
  \item Intervene on the specified variable; keep other facts fixed.
  \item Recompute downstream effects; allow alternate pathways.
  \item Output \texttt{yes}/\texttt{no} based on whether the outcome occurs.
\end{enumerate}
\\
\midrule

LLM Paragraph &
\{
"summary": "Trace the causal structure (AND/OR). Use observations to fix actual states. In counterfactuals, change only the intervened variable. Recompute downstream deterministically.",
"compact\_summary": "Trace causation; use observations; isolate interventions; recompute downstream."
\}
\\

\bottomrule
\end{tabularx}
\caption{Example summaries produced by different summary prompts (one example per prompt).}
\label{tab:summary_examples}
\end{table}

\subsection{Error Selection}
\begin{table}[H]
\centering
\small
\setlength{\tabcolsep}{4pt}
\renewcommand{\arraystretch}{1.05}
\begin{tabular*}{\textwidth}{
@{\extracolsep{\fill}}
l
c c        % CounterBench
c            % spacer
c c        % GSM8K
c            % spacer
c c        % ReasonIF
@{}
}
\toprule
& \multicolumn{2}{c}{\textbf{CounterBench (\%)}} &
  \multicolumn{1}{@{}c@{}}{}{}{}{}{\hspace{0.9em}} &
  \multicolumn{2}{c}{\textbf{GSM8K (\%)}} &
  \multicolumn{1}{@{}c@{}}{}{}{}{}{\hspace{0.9em}} &
  \multicolumn{2}{c}{\textbf{ReasonIF (\%)}} \\
\cmidrule(lr){2-3}
\cmidrule(lr){5-6}
\cmidrule(lr){8-9}
\textbf{Model} 
& $k$-means & Random
& \multicolumn{1}{@{}c@{}}{}{}{}{}{}
& $k$-means & Random
& \multicolumn{1}{@{}c@{}}{}{}{}{}{}
& $k$-means & Random  \\
\midrule
\rowcolor{lightlightgray}
\multicolumn{9}{c}{\textbf{\textit{Gemma-3 Instruct Models}}} \\
Gemma-1B  & 51.6 & \textbf{51.9}
& \multicolumn{1}{@{}c@{}}{}{}{}{}{}
& \textbf{46.0} & 43.0
& \multicolumn{1}{@{}c@{}}{}{}{}{}{}
& \textbf{50.0} & 40.0 \\

Gemma-4B  & \textbf{74.3} & 71.3
& \multicolumn{1}{@{}c@{}}{}{}{}{}{}
& \textbf{86.1} & 83.6
& \multicolumn{1}{@{}c@{}}{}{}{}{}{}
& \textbf{66.0} & 60.3 \\

Gemma-12B & 84.9 & 80.4
& \multicolumn{1}{@{}c@{}}{}{}{}{}{}
& \textbf{93.8} & 92.8
& \multicolumn{1}{@{}c@{}}{}{}{}{}{}
& \textbf{74.0} & 69.3 \\

Gemma-27B & 80.7 & \textbf{80.8}
& \multicolumn{1}{@{}c@{}}{}{}{}{}{}
& \textbf{95.5} & 95.0
& \multicolumn{1}{@{}c@{}}{}{}{}{}{}
& \textbf{76.7} & 64.3 \\

\midrule
\rowcolor{lightlightgray}
\multicolumn{9}{c}{\textbf{\textit{Qwen-2.5 Instruct Models}}} \\
Qwen-0.5B & \textbf{42.6} & 24.0
& \multicolumn{1}{@{}c@{}}{}{}{}{}{}
& \textbf{24.9} & 1.4 
& \multicolumn{1}{@{}c@{}}{}{}{}{}{}
& \textbf{35.3} & \textbf{35.3} \\

Qwen-1.5B & \textbf{47.8} & 39.2
& \multicolumn{1}{@{}c@{}}{}{}{}{}{}
& \textbf{53.4} & 13.8
& \multicolumn{1}{@{}c@{}}{}{}{}{}{}
& \textbf{35.0} & 32.0 \\

Qwen-3B   & \textbf{58.1} & 52.9
& \multicolumn{1}{@{}c@{}}{}{}{}{}{}
& \textbf{84.8} & 77.3
& \multicolumn{1}{@{}c@{}}{}{}{}{}{}
& \textbf{42.3} & 41.3 \\

Qwen-7B   & 78.0 & \textbf{78.1}
& \multicolumn{1}{@{}c@{}}{}{}{}{}{}
& \textbf{91.0} & 90.4
& \multicolumn{1}{@{}c@{}}{}{}{}{}{}
& \textbf{67.7} & 53.0 \\

Qwen-14B  & \textbf{80.6} & 73.3
& \multicolumn{1}{@{}c@{}}{}{}{}{}{}
& \textbf{94.8} & 94.1
& \multicolumn{1}{@{}c@{}}{}{}{}{}{}
& \textbf{95.0} & 74.3 \\

Qwen-32B  & \textbf{84.6} & 78.4
& \multicolumn{1}{@{}c@{}}{}{}{}{}{}
& \textbf{96.4} & 96.1
& \multicolumn{1}{@{}c@{}}{}{}{}{}{}
& \textbf{96.3} & 83.7 \\

Qwen-72B  & 88.9 & 84.6
& \multicolumn{1}{@{}c@{}}{}{}{}{}{}
& 95.7 & \textbf{95.8}
& \multicolumn{1}{@{}c@{}}{}{}{}{}{}
& \textbf{93.7} & 75.0 \\

\bottomrule
\end{tabular*}
\caption{Full ablation: error selection strategy.}
\label{tab:ablation_error_selection_full}
\end{table}

\begin{table}[H]
\centering
\small
\setlength{\tabcolsep}{3pt}
\renewcommand{\arraystretch}{1.05}
\begin{tabular*}{\textwidth}{
@{\extracolsep{\fill}}
l
c c c   % k=1 + spacer
c c c   % k=2 + spacer
c c c   % k=3 + spacer
c c c   % k=4 + spacer
c c c   % k=5 + spacer
c c     % k=6 (no trailing spacer)
@{}
}
\toprule
& \multicolumn{2}{c}{$k=1$} &
& \multicolumn{2}{c}{$k=2$} &
& \multicolumn{2}{c}{$k=3$} &
& \multicolumn{2}{c}{$k=4$} &
& \multicolumn{2}{c}{$k=5$} &
& \multicolumn{2}{c}{$k=6$} \\

\cmidrule(lr){2-3}
\cmidrule(lr){5-6}
\cmidrule(lr){8-9}
\cmidrule(lr){11-12}
\cmidrule(lr){14-15}
\cmidrule(lr){17-18}
\textbf{Model} 

& $k$-means & TT &
& $k$-means & TT &
& $k$-means & TT &
& $k$-means & TT &
& $k$-means & TT &
& $k$-means & TT \\
\midrule

\rowcolor{lightlightgray}
\multicolumn{18}{c}{\textbf{\textit{Gemma-3 Instruct Models}}} \\

Gemma-1B
& \textbf{40.3} & 25.3 & {}
& \textbf{46.3} & 45.7 & {}
& \textbf{49.7} & 46.3 & {}
& \textbf{36.7} & 29.7 & {}
& \textbf{44.0} & 31.7 & {}
& \textbf{39.3} & 37.3 \\

Gemma-4B
& \textbf{64.0} & 49.7 & {}
& \textbf{54.0} & 39.0 & {}
& \textbf{59.0} & 56.0 & {}
& \textbf{64.7} & 56.3 & {}
& \textbf{61.3} & 50.3 & {}
& \textbf{61.3} & 52.7 \\

Gemma-12B
& 64.7 & \textbf{66.3} & {}
& \textbf{77.0} & 34.7 & {}
& \textbf{68.0} & 53.0 & {}
& \textbf{70.3} & 65.7 & {}
& \textbf{65.0} & 61.0 & {}
& \textbf{70.7} & 67.0 \\

Gemma-27B
& \textbf{68.7} & 60.3 & {}
& \textbf{76.7} & 25.7 & {}
& \textbf{85.3} & 29.3 & {}
& \textbf{82.0} & 76.7 & {}
& 78.3 & \textbf{78.3} & {}
& \textbf{76.3} & 75.3 \\

\midrule
\rowcolor{lightlightgray}
\multicolumn{18}{c}{\textbf{\textit{Qwen-2.5 Instruct Models}}} \\

Qwen-0.5B
& 32.7 & \textbf{33.0} & {}
& 27.7 & \textbf{28.7} & {}
& \textbf{36.7} & 27.3 & {}
& 25.7 & \textbf{42.0} & {}
& \textbf{35.3} & 31.3 & {}
& \textbf{38.0} & 32.7 \\

Qwen-1.5B
& 29.3 & \textbf{38.3} & {}
& 31.3 & \textbf{39.7} & {}
& 28.7 & \textbf{32.0} & {}
& 28.3 & \textbf{34.7} & {}
& \textbf{30.7} & 29.3 & {}
& \textbf{35.3} & 25.7 \\

Qwen-3B
& \textbf{28.3} & 18.7 & {}
& \textbf{56.0} & 24.0 & {}
& 37.7 & \textbf{56.3} & {}
& 27.3 & \textbf{51.0} & {}
& \textbf{50.3} & 29.0 & {}
& \textbf{49.0} & 7.7 \\

Qwen-7B
& \textbf{61.0} & 57.7 & {}
& \textbf{54.0} & 37.7 & {}
& \textbf{41.0} & 32.3 & {}
& \textbf{64.3} & 31.3 & {}
& 49.0 & \textbf{62.3} & {}
& \textbf{52.3} & 43.7 \\

Qwen-14B
& 76.0 & \textbf{86.0} & {}
& \textbf{78.3} & 74.0 & {}
& \textbf{86.3} & 67.0 & {}
& 75.3 & \textbf{80.7} & {}
& \textbf{88.3} & 77.7 & {}
& 79.7 & \textbf{81.7} \\

Qwen-32B
& \textbf{81.3} & 65.7 & {}
& 82.0 & \textbf{83.7} & {}
& \textbf{94.0} & 72.7 & {}
& \textbf{87.3} & 64.3 & {}
& 90.7 & \textbf{91.3} & {}
& \textbf{90.7} & 75.0 \\

Qwen-72B
& 78.3 & \textbf{87.7} & {}
& \textbf{90.3} & 23.3 & {}
& \textbf{84.3} & 84.0 & {}
& \textbf{87.7} & 77.3 & {}
& \textbf{92.0} & 46.7 & {}
& \textbf{92.3} & 48.3 \\

\bottomrule
\end{tabular*}
\caption{ReasonIF performance vs.\ $k$, comparing $k$-means clustering with task-type selection (TT).}
\label{tab:ablation_k_sweep_full}
\end{table}

\FloatBarrier
\subsection{Explanation Quality}
\begin{table}[H]
\centering
\small
\setlength{\tabcolsep}{4pt}
\renewcommand{\arraystretch}{1.05}
\begin{tabular*}{\textwidth}{
@{\extracolsep{\fill}}
l
c c c        % CounterBench
c            % spacer
c c c        % GSM8K
c            % spacer
c c c        % ReasonIF
@{}
}
\toprule
& \multicolumn{3}{c}{\textbf{CounterBench (\%)}} &
  \multicolumn{1}{@{}c@{}}{}{}{}{\hspace{0.9em}} &
  \multicolumn{3}{c}{\textbf{GSM8K (\%)}} &
  \multicolumn{1}{@{}c@{}}{}{}{}{\hspace{0.9em}} &
  \multicolumn{3}{c}{\textbf{ReasonIF (\%)}} \\
\cmidrule(lr){2-4}
\cmidrule(lr){6-8}
\cmidrule(lr){10-12}
\textbf{Model}
& Verified & Unverified & Solution
& \multicolumn{1}{@{}c@{}}{}{}{}{}
& Verified & Unverified & Solution
& \multicolumn{1}{@{}c@{}}{}{}{}{}
& Verified & Unverified & Solution \\
\midrule
\rowcolor{lightlightgray}
\multicolumn{12}{c}{\textbf{\textit{Gemma-3 Instruct Models}}} \\
Gemma-1B  & \textbf{51.6} & 47.0 & 48.8
& \multicolumn{1}{@{}c@{}}{}{}{}{}
& \textbf{46.0} & 40.0 & 44.3
& \multicolumn{1}{@{}c@{}}{}{}{}{}
& \textbf{50.0} & 40.7 & 37.0 \\

Gemma-4B  & \textbf{74.3} & 61.7 & 65.9
& \multicolumn{1}{@{}c@{}}{}{}{}{}
& \textbf{86.1} & 84.5 & 85.3
& \multicolumn{1}{@{}c@{}}{}{}{}{}
& \textbf{66.0} & 54.0 & 35.7 \\

Gemma-12B & \textbf{84.9} & 75.1 & 77.0
& \multicolumn{1}{@{}c@{}}{}{}{}{}
& \textbf{93.8} & 91.4 & 93.4
& \multicolumn{1}{@{}c@{}}{}{}{}{}
& \textbf{74.0} & 61.3 & 61.7 \\

Gemma-27B & \textbf{80.7} & 78.6 & 74.6
& \multicolumn{1}{@{}c@{}}{}{}{}{}
& \textbf{95.5} & 95.3 & 95.1
& \multicolumn{1}{@{}c@{}}{}{}{}{}
& \textbf{76.7} & 77.3 & 76.0 \\

\midrule
\rowcolor{lightlightgray}
\multicolumn{12}{c}{\textbf{\textit{Qwen-2.5 Instruct Models}}} \\
Qwen-0.5B & \textbf{42.6} &  9.2 & 12.4
& \multicolumn{1}{@{}c@{}}{}{}{}{}
& \textbf{24.9} &  0.9 & 19.9
& \multicolumn{1}{@{}c@{}}{}{}{}{}
& \textbf{35.3} & 30.3 & 33.0 \\

Qwen-1.5B & \textbf{47.8} & 10.3 & 23.6
& \multicolumn{1}{@{}c@{}}{}{}{}{}
& \textbf{53.4} & 11.8 & 36.2
& \multicolumn{1}{@{}c@{}}{}{}{}{}
& 35.0 & \textbf{35.7} & 30.7 \\

Qwen-3B   & 58.1 & 48.8 & \textbf{58.6}
& \multicolumn{1}{@{}c@{}}{}{}{}{}
& \textbf{84.8} & 76.0 & 81.7
& \multicolumn{1}{@{}c@{}}{}{}{}{}
& \textbf{42.3} & 35.3 & 39.0 \\

Qwen-7B   & \textbf{78.0} & 74.3 & 72.9
& \multicolumn{1}{@{}c@{}}{}{}{}{}
& \textbf{91.0} & 87.4 & 88.9
& \multicolumn{1}{@{}c@{}}{}{}{}{}
& \textbf{67.7} & 38.3 & 40.0 \\

Qwen-14B  & \textbf{80.6} & 74.9 & 75.9
& \multicolumn{1}{@{}c@{}}{}{}{}{}
& \textbf{94.8} & 94.5 & 93.9
& \multicolumn{1}{@{}c@{}}{}{}{}{}
& \textbf{95.0} & 71.0 & 84.0 \\

Qwen-32B  & \textbf{84.6} & 83.2 & 83.2
& \multicolumn{1}{@{}c@{}}{}{}{}{}
& \textbf{96.4} & 83.3 & 90.4
& \multicolumn{1}{@{}c@{}}{}{}{}{}
& \textbf{96.3} & 69.0 & 76.3 \\

Qwen-72B  & \textbf{88.9} & 86.4 & 87.7
& \multicolumn{1}{@{}c@{}}{}{}{}{}
& \textbf{95.7} & 88.9 & 93.1
& \multicolumn{1}{@{}c@{}}{}{}{}{}
& \textbf{93.7} & 81.7 & 84.0 \\

\bottomrule
\end{tabular*}
\caption{Full ablation: explanatory explanation verification.}
\label{tab:ablation_feedback_quality_full}
\end{table}

\FloatBarrier
\subsection{Summarization vs.\ Raw Explanations}

\begin{table}[H]
\centering
\small
\setlength{\tabcolsep}{4pt}
\renewcommand{\arraystretch}{1.05}
\begin{tabular*}{\textwidth}{
@{\extracolsep{\fill}}
l
c c        % CounterBench
c            % spacer
c c        % GSM8K
c            % spacer
c c        % ReasonIF
@{}
}
\toprule
& \multicolumn{2}{c}{\textbf{CounterBench (\%)}} &
  \multicolumn{1}{@{}c@{}}{}{}{\hspace{0.9em}} &
  \multicolumn{2}{c}{\textbf{GSM8K (\%)}} &
  \multicolumn{1}{@{}c@{}}{}{}{\hspace{0.9em}} &
  \multicolumn{2}{c}{\textbf{ReasonIF (\%)}} \\
\cmidrule(lr){2-3}
\cmidrule(lr){5-6}
\cmidrule(lr){8-9}
\textbf{Model}
& Summary & Raw Expl.
& \multicolumn{1}{@{}c@{}}{}{}{}
& Summary & Raw Expl.
& \multicolumn{1}{@{}c@{}}{}{}{}
& Summary & Raw Expl.  \\
\midrule

\rowcolor{lightlightgray}
\multicolumn{9}{c}{\textbf{\textit{Gemma-3 Instruct Models}}} \\
Gemma-1B  & \textbf{51.6} & 44.4
& \multicolumn{1}{@{}c@{}}{}{}{}
& \textbf{46.0} & 43.6
& \multicolumn{1}{@{}c@{}}{}{}{}
& \textbf{50.0} & 32.7 \\

Gemma-4B  & \textbf{74.3} & 71.7
& \multicolumn{1}{@{}c@{}}{}{}{}
& \textbf{86.1} & 85.6
& \multicolumn{1}{@{}c@{}}{}{}{}
& \textbf{66.0} & 56.3 \\

Gemma-12B & \textbf{84.9} & 78.7
& \multicolumn{1}{@{}c@{}}{}{}{}
& \textbf{93.8} & 93.0
& \multicolumn{1}{@{}c@{}}{}{}{}
& \textbf{74.0} & 49.0 \\

Gemma-27B & 80.7 & \textbf{81.8}
& \multicolumn{1}{@{}c@{}}{}{}{}
& \textbf{95.5} & 95.1
& \multicolumn{1}{@{}c@{}}{}{}{}
& \textbf{76.7} & 45.7 \\

\midrule
\rowcolor{lightlightgray}
\multicolumn{9}{c}{\textbf{\textit{Qwen-2.5 Instruct Models}}} \\

Qwen-0.5B & \textbf{42.6} & 21.0
& \multicolumn{1}{@{}c@{}}{}{}{}
& \textbf{24.9} & 1.3 
& \multicolumn{1}{@{}c@{}}{}{}{}
& \textbf{35.3} & 19.7 \\

Qwen-1.5B & \textbf{47.8} & 39.9
& \multicolumn{1}{@{}c@{}}{}{}{}
& \textbf{53.4} & 18.3
& \multicolumn{1}{@{}c@{}}{}{}{}
& 35.0 & \textbf{37.0} \\

Qwen-3B   & 58.1 & \textbf{62.7}
& \multicolumn{1}{@{}c@{}}{}{}{}
& \textbf{84.8} & 67.6
& \multicolumn{1}{@{}c@{}}{}{}{}
& \textbf{42.3} & 32.3 \\

Qwen-7B   & \textbf{78.0} & 71.5
& \multicolumn{1}{@{}c@{}}{}{}{}
& \textbf{91.0} & 90.4
& \multicolumn{1}{@{}c@{}}{}{}{}
& \textbf{67.7} & 66.8 \\

Qwen-14B  & \textbf{80.6} & 72.9
& \multicolumn{1}{@{}c@{}}{}{}{}
& \textbf{94.8} & 93.9
& \multicolumn{1}{@{}c@{}}{}{}{}
& \textbf{95.0} & 67.7 \\

Qwen-32B  & \textbf{84.6} & 79.2
& \multicolumn{1}{@{}c@{}}{}{}{}
& \textbf{96.4} & 96.1
& \multicolumn{1}{@{}c@{}}{}{}{}
& \textbf{96.3} & 87.3 \\

Qwen-72B  & \textbf{88.9} & 77.5
& \multicolumn{1}{@{}c@{}}{}{}{}
& \textbf{95.7} & 94.9
& \multicolumn{1}{@{}c@{}}{}{}{}
& \textbf{93.7} & 65.0 \\

\bottomrule
\end{tabular*}
\caption{Full ablation: summarization vs raw concatenated explanations.}
\label{tab:ablation_summarization_full}
\end{table}

\FloatBarrier
\subsection{Summary Selection and Ranking}

\begin{table}[H]
\centering
\small
\setlength{\tabcolsep}{4pt}
\renewcommand{\arraystretch}{1.05}
\begin{tabular*}{\textwidth}{
@{\extracolsep{\fill}}
l
c c c        % CounterBench
c            % spacer
c c c        % GSM8K
c            % spacer
c c c        % ReasonIF
@{}
}
\toprule
& \multicolumn{3}{c}{\textbf{CounterBench (\%)}} &
  \multicolumn{1}{@{}c@{}}{}{}{}{}{}{\hspace{0.9em}} &
  \multicolumn{3}{c}{\textbf{GSM8K (\%)}} &
  \multicolumn{1}{@{}c@{}}{}{}{}{}{}{\hspace{0.9em}} &
  \multicolumn{3}{c}{\textbf{ReasonIF (\%)}} \\
\cmidrule(lr){2-4}
\cmidrule(lr){6-8}
\cmidrule(lr){10-12}
\textbf{Model}
& Best & Median & Worst
& \multicolumn{1}{@{}c@{}}{}{}{}{}{}{}
& Best & Median & Worst
& \multicolumn{1}{@{}c@{}}{}{}{}{}{}{}
& Best & Median & Worst \\

\midrule

\rowcolor{lightlightgray}
\multicolumn{12}{c}{\textbf{\textit{Gemma-3 Instruct Models}}} \\
Gemma-1B  & \textbf{51.6} & 49.7 & 49.9
& \multicolumn{1}{@{}c@{}}{}{}{}{}{}{}
& \textbf{46.0} & 38.7 & 43.1 
& \multicolumn{1}{@{}c@{}}{}{}{}{}{}{}
& \textbf{50.0} & 42.7 & 41.7 \\

Gemma-4B  & \textbf{74.3} & 70.2 & 68.4
& \multicolumn{1}{@{}c@{}}{}{}{}{}{}{}
& \textbf{86.1} & 84.8 & 85.4 
& \multicolumn{1}{@{}c@{}}{}{}{}{}{}{}
& 66.0 & 65.7 & \textbf{69.0} \\

Gemma-12B & \textbf{84.9} & 70.0 & 72.3
& \multicolumn{1}{@{}c@{}}{}{}{}{}{}{}
& \textbf{93.8} & 93.1 & 92.8 
& \multicolumn{1}{@{}c@{}}{}{}{}{}{}{}
& \textbf{74.0} & 72.7 & 71.0 \\

Gemma-27B & \textbf{80.7} & 78.3 & 73.2
& \multicolumn{1}{@{}c@{}}{}{}{}{}{}{}
& \textbf{95.5} & 94.9 & 94.9 
& \multicolumn{1}{@{}c@{}}{}{}{}{}{}{}
& \textbf{76.7} & 74.0 & 70.7 \\

\midrule
\rowcolor{lightlightgray}
\multicolumn{12}{c}{\textbf{\textit{Qwen-2.5 Instruct Models}}} \\

Qwen-0.5B & \textbf{42.6} & 17.5 & 15.5
& \multicolumn{1}{@{}c@{}}{}{}{}{}{}{}
& \textbf{24.9}  & 23.0  & 13.3  
& \multicolumn{1}{@{}c@{}}{}{}{}{}{}{}
& \textbf{35.3} & 30.0 & 23.7 \\

Qwen-1.5B & \textbf{47.8} & 45.7 & 33.3
& \multicolumn{1}{@{}c@{}}{}{}{}{}{}{}
& \textbf{53.4} & 45.3 & 33.1 
& \multicolumn{1}{@{}c@{}}{}{}{}{}{}{}
& \textbf{35.0} & 31.0 & 24.3 \\

Qwen-3B   & \textbf{58.1} & 54.1 & 47.9
& \multicolumn{1}{@{}c@{}}{}{}{}{}{}{}
& \textbf{84.8} & 83.9 & 82.8 
& \multicolumn{1}{@{}c@{}}{}{}{}{}{}{}
& \textbf{42.3} & 30.7 & 22.7 \\

Qwen-7B   & \textbf{78.0} & 74.8 & 74.2
& \multicolumn{1}{@{}c@{}}{}{}{}{}{}{}
& 91.0 & 90.4 & \textbf{91.7} 
& \multicolumn{1}{@{}c@{}}{}{}{}{}{}{}
& \textbf{67.7} & 56.0 & 50.7 \\

Qwen-14B  & \textbf{80.6} & 77.2 & 77.7
& \multicolumn{1}{@{}c@{}}{}{}{}{}{}{}
& \textbf{94.8} & 93.3 & 93.8 
& \multicolumn{1}{@{}c@{}}{}{}{}{}{}{}
& \textbf{95.0} & 86.0 & 81.3 \\

Qwen-32B  & \textbf{84.6} & 83.5 & 79.2
& \multicolumn{1}{@{}c@{}}{}{}{}{}{}{}
& \textbf{96.4} & 96.1 & 95.4 
& \multicolumn{1}{@{}c@{}}{}{}{}{}{}{}
& \textbf{96.3} & 92.7 & 88.3 \\

Qwen-72B  & \textbf{88.9} & 86.7 & 83.8
& \multicolumn{1}{@{}c@{}}{}{}{}{}{}{}
& \textbf{95.7} & 95.7 & 95.6 
& \multicolumn{1}{@{}c@{}}{}{}{}{}{}{}
& \textbf{93.7} & 82.3 & 79.7 \\

\bottomrule
\end{tabular*}
\caption{Full ablation: summary ranking by $\Delta$-embedding score. }
\label{tab:ablation_summary_rank_full}
\end{table}

\begin{table}[H]
\centering
\small
\setlength{\tabcolsep}{4pt}
\renewcommand{\arraystretch}{1.05}
\begin{tabular*}{\textwidth}{
@{\extracolsep{\fill}}
l
c c        % CounterBench
c            % spacer
c c        % GSM8K
c            % spacer
c c        % ReasonIF
@{}
}\toprule
& \multicolumn{2}{c}{\textbf{CounterBench (\%)}} &
  \multicolumn{1}{@{}c@{}}{}{\hspace{0.9em}} &
  \multicolumn{2}{c}{\textbf{GSM8K (\%)}} &
   \multicolumn{1}{@{}c@{}}{}{\hspace{0.9em}} &
  \multicolumn{2}{c}{\textbf{ReasonIF (\%)}} \\
\cmidrule(lr){2-3}
\cmidrule(lr){5-6}
\cmidrule(lr){8-9}
\textbf{Model}
& Weighted &  Unweighted
& \multicolumn{1}{@{}c@{}}{}{}
& Weighted &  Unweighted
& \multicolumn{1}{@{}c@{}}{}{}
& Weighted &  Unweighted \\
\midrule

\rowcolor{lightlightgray}
\multicolumn{9}{c}{\textbf{\textit{Gemma-3 Instruct Models}}} \\
Gemma-1B  & \textbf{51.6} & 49.0
& \multicolumn{1}{@{}c@{}}{}{} & \textbf{46.0} & 40.4
& \multicolumn{1}{@{}c@{}}{}{} & \textbf{50.0} & 31.3 \\

Gemma-4B  & \textbf{74.3} & 72.7
& \multicolumn{1}{@{}c@{}}{}{} & \textbf{86.1} & 85.1
& \multicolumn{1}{@{}c@{}}{}{} & 66.0 & \textbf{69.0} \\

Gemma-12B & \textbf{84.9} & 70.0
& \multicolumn{1}{@{}c@{}}{}{} & \textbf{93.8} & 92.9
& \multicolumn{1}{@{}c@{}}{}{} & \textbf{74.0} & 72.7 \\

Gemma-27B & \textbf{80.7} & 78.3
& \multicolumn{1}{@{}c@{}}{}{} & \textbf{95.5} & 94.4
& \multicolumn{1}{@{}c@{}}{}{} & 76.7 & \textbf{80.0} \\

\midrule
\rowcolor{lightlightgray}
\multicolumn{9}{c}{\textbf{\textit{Qwen-2.5 Instruct Models}}} \\

Qwen-0.5B & \textbf{42.6} & 31.8
& \multicolumn{1}{@{}c@{}}{}{}
& \textbf{24.9} & 22.2 
& \multicolumn{1}{@{}c@{}}{}{}
& \textbf{35.3} & 35.3 \\

Qwen-1.5B & \textbf{47.8} & 31.6
& \multicolumn{1}{@{}c@{}}{}{}
& \textbf{53.4} & 53.4
& \multicolumn{1}{@{}c@{}}{}{}
& \textbf{35.0} & 35.0 \\

Qwen-3B   & \textbf{58.1} & 58.1
& \multicolumn{1}{@{}c@{}}{}{}
& \textbf{84.8} & 81.0
& \multicolumn{1}{@{}c@{}}{}{}
& 42.3 & \textbf{45.7} \\

Qwen-7B   & 78.0 & \textbf{78.9}
& \multicolumn{1}{@{}c@{}}{}{}
& \textbf{91.0} & 91.0
& \multicolumn{1}{@{}c@{}}{}{}
& \textbf{67.7} & 62.3 \\

Qwen-14B  & \textbf{80.6} & 80.6
& \multicolumn{1}{@{}c@{}}{}{}
& \textbf{94.8} & 92.8
& \multicolumn{1}{@{}c@{}}{}{}
& \textbf{95.0} & 86.0 \\

Qwen-32B  & \textbf{84.6} & 82.7
& \multicolumn{1}{@{}c@{}}{}{}
& \textbf{96.4} & 96.4
& \multicolumn{1}{@{}c@{}}{}{}
& \textbf{96.3} & 89.0 \\

Qwen-72B  & \textbf{88.9} & 82.1
& \multicolumn{1}{@{}c@{}}{}{}
& \textbf{95.7} & 95.7
& \multicolumn{1}{@{}c@{}}{}{}
& \textbf{93.7} & 71.0 \\

\bottomrule
\end{tabular*}
\caption{Full ablation: weighted vs unweighted summary scoring.}
\label{tab:ablation_weighted_full}
\end{table}

\FloatBarrier

\subsection{FLEx Combined with Other Methods}

\begin{table}[H]
\centering
\scriptsize
\setlength{\tabcolsep}{2.6pt}
\renewcommand{\arraystretch}{1.05}
\resizebox{\textwidth}{!}{
\begin{tabular}{
l
c c c c c c c
c
c c c c c c c
c
c c c c c c c
}
\toprule
% Dataset row
& \multicolumn{7}{c}{\textbf{CounterBench (\%)}} &
& \multicolumn{7}{c}{\textbf{GSM8K (\%)}} &
& \multicolumn{7}{c}{\textbf{ReasonIF (\%)}} \\
\cmidrule(lr){2-8}
\cmidrule(lr){10-16}
\cmidrule(lr){18-24}

% Method group row
% \multirow{2}{*}{\textbf{Model}}
& \multirow{2}{*}{\textbf{FLEx}}
& \multicolumn{2}{c}{SR}
& \multicolumn{2}{c}{RAG}
& \multicolumn{2}{c}{SC}
&
& \multirow{2}{*}{\textbf{FLEx}}
& \multicolumn{2}{c}{SR}
& \multicolumn{2}{c}{RAG}
& \multicolumn{2}{c}{SC}
&
& \multirow{2}{*}{\textbf{FLEx}}
& \multicolumn{2}{c}{SR}
& \multicolumn{2}{c}{RAG}
& \multicolumn{2}{c}{SC} \\
\cmidrule(lr){3-4}\cmidrule(lr){5-6}\cmidrule(lr){7-8}
\cmidrule(lr){11-12}\cmidrule(lr){13-14}\cmidrule(lr){15-16}
\cmidrule(lr){19-20}\cmidrule(lr){21-22}\cmidrule(lr){23-24}
% Base / w FLEx row (Model goes here)
\textbf{Model}
% Base / w FLEx row
& 
& base & w/ FLEx
& base & w/ FLEx
& base & w/ FLEx
&
&
& base & w/ FLEx
& base & w/ FLEx
& base & w/ FLEx
&
&
& base & w/ FLEx
& base & w/ FLEx
& base & w/ FLEx \\
\midrule

\rowcolor{lightlightgray}
\multicolumn{24}{c}{\textbf{\textit{Gemma-3 Instruct Models}}} \\

Gemma-1B
& 51.6 & 49.7 & \textbf{51.9} & 48.5 & 49.0 & 51.1 & 49.3
& &
  46.0 & 44.8 & 43.7 & 38.0 & 37.7 & \textbf{53.2} & 49.6
& &
  \textbf{50.0} & 38.0 & \textbf{50.0} & 46.3 & 42.0 & 36.3 & 49.7 \\

Gemma-4B
& 74.3 & 65.4 & 71.8 & 69.7 & 70.6 & 68.5 & \textbf{77.0}
& &
  86.1 & 84.8 & 84.9 & 85.2 & 84.2 & 87.7 & \textbf{87.9}
& &
  \textbf{66.0} & 47.3 & 63.3 & 58.0 & 56.0 & 46.7 & 63.3 \\

Gemma-12B
& 84.9 & 66.7 & 79.2 & 70.4 & 81.3 & 72.1 & \textbf{87.0}
& &
  93.8 & 93.1 & 93.1 & 92.9 & 92.2 & 94.0 & \textbf{94.3}
& &
  \textbf{74.0} & 58.0 & 63.7 & 71.0 & 68.0 & 63.3 & 71.7 \\

Gemma-27B
& 80.7 & 74.6 & 80.8 & 75.9 & 80.2 & 78.2 & \textbf{81.5}
& &
  95.5 & 94.7 & 94.8 & 94.5 & 94.2 & 95.1 & \textbf{95.8}
& &
  76.7 & 68.7 & 73.0 & 77.3 & \textbf{81.0} & 73.7 & 78.0 \\

\midrule
\rowcolor{lightlightgray}
\multicolumn{24}{c}{\textbf{\textit{Qwen-2.5 Instruct Models}}} \\

Qwen-0.5B
& 42.6 & 21.8 & 42.7 & \textbf{43.4} & 40.8 & 38.0 & 42.0
& &
  24.9 & 29.0 & 29.4 & 40.2 & \textbf{42.0} & 34.3 & 32.7
& &
  35.3 & 35.7 & 42.3 & 45.0 & \textbf{50.0} & 29.0 & 36.7 \\

Qwen-1.5B
& 47.8 & 47.0 & 49.1 & 49.7 & 49.5 & 47.8 & \textbf{54.2}
& &
  53.4 & 51.8 & 54.1 & 69.4 & \textbf{71.2} & 59.0 & 63.2
& &
  35.0 & 36.3 & 38.0 & \textbf{48.0} & 45.0 & 31.3 & 34.0 \\

Qwen-3B
& 58.1 & 53.3 & 57.7 & \textbf{61.3} & 60.3 & 55.2 & 60.2
& &
  84.8 & 84.0 & 84.7 & 85.6 & 84.8 & 87.6 & \textbf{89.0}
& &
  42.3 & 44.7 & 46.3 & 38.7 & \textbf{49.3} & 39.3 & 43.3 \\

Qwen-7B
& 78.0 & 68.8 & 75.4 & 68.5 & 74.3 & 69.0 & \textbf{80.3}
& &
  91.0 & 88.2 & 89.1 & 90.7 & 91.4 & \textbf{93.0} & 92.7
& &
  67.7 & 56.7 & 63.7 & 66.7 & 60.7 & 47.3 & \textbf{70.0} \\

Qwen-14B
& 80.6 & 74.7 & \textbf{81.3} & 78.3 & 80.7 & 75.8 & 80.2
& &
  94.8 & 91.7 & 94.3 & 94.4 & 93.8 & 95.5 & \textbf{95.6}
& &
  \textbf{95.0} & 70.0 & 91.3 & 75.3 & 78.3 & 72.3 & 94.0 \\

Qwen-32B
& 84.6 & 80.3 & 81.8 & \textbf{85.9} & \textbf{85.9} & 80.2 & 84.8
& &
  \textbf{96.4} & 94.1 & 96.3 & 94.8 & 94.9 & 92.5 & 96.3
& &
  \textbf{96.3} & 75.0 & 85.3 & 79.7 & 74.0 & 79.0 & 95.3 \\

Qwen-72B
& 88.9 & 81.3 & 87.3 & 86.5 & 89.1 & 85.2 & \textbf{90.9}
& &
  95.7 & 94.8 & 95.6 & 95.1 & 95.5 & 93.6 & \textbf{96.3}
& &
  93.7 & 85.3 & 94.0 & 83.7 & 85.7 & 83.0 & \textbf{96.0} \\

\bottomrule
\end{tabular}
}
\caption{
Performance (\% accuracy) on self-refine (SR), retrieval-augmented generation (RAG), and self-consistency (SC). For each method, we report the base method alone and when augmented with FLEx. Bold indicates the best result per model and dataset.
}
\label{tab:flex_plus_others}
\end{table}

\subsection{Summary Overhead \& Generalization }

\FloatBarrier

\begin{table}[H]
\centering
\small
\setlength{\tabcolsep}{6pt}
\renewcommand{\arraystretch}{1.1}
\begin{tabular*}{\textwidth}{@{\extracolsep{\fill}} l c c c @{}}
\toprule
\textbf{Model} &
\textbf{CounterBench (tokens)} &
\textbf{GSM8K (tokens)} &
\textbf{ReasonIF (tokens)} \\
\midrule

\rowcolor{lightlightgray}
\multicolumn{4}{c}{\textbf{\textit{Gemma-3 Instruct Models}}} \\
Gemma-1B  & 234 & 126 & 250 \\
Gemma-4B  & 102 & 301 & 122 \\
Gemma-12B & 247 & 173 & 111 \\
Gemma-27B & 119 & 115 & 300 \\

\midrule
\rowcolor{lightlightgray}
\multicolumn{4}{c}{\textbf{\textit{Qwen-2.5 Instruct Models}}} \\
Qwen-0.5B & 87  & 84  & 94  \\
Qwen-1.5B & 90  & 73  & 119 \\
Qwen-3B   & 203 & 73  & 245 \\
Qwen-7B   & 168 & 222 & 139 \\
Qwen-14B  & 268 & 287 & 134 \\
Qwen-32B  & 100 & 158 & 96  \\
Qwen-72B  & 105 & 247 & 158 \\
\midrule
Average & 157 & 169 & 161 \\
\bottomrule
\end{tabular*}
\caption{Overhead of FLEx, measured as additional prompt tokens introduced by the learned summary.}
\label{tab:inference_overhead_tokens}
\end{table}

\begin{table}[H]
\centering
\small
\setlength{\tabcolsep}{6pt}
\renewcommand{\arraystretch}{1.1}
\begin{tabular*}{\textwidth}{@{\extracolsep{\fill}} l c c c @{}}
\toprule
\textbf{Source Model} &
\textbf{CounterBench} &
\textbf{GSM8K} &
\textbf{ReasonIF} \\
\midrule

\rowcolor{lightlightgray}
\multicolumn{4}{c}{\textbf{\textit{Gemma-3 Instruct Models}}} \\
Gemma-1B        & $\uparrow$ 0.9 & $\downarrow$ 0.4 & $\downarrow$ 15.3 \\
Gemma-4B        & $\downarrow$ 0.1 & $\downarrow$ 0.9 & $\downarrow$ 16.6 \\
Gemma-12B       & $\uparrow$ 4.1 & $\uparrow$ 0.2 & $\downarrow$ 21.3 \\
Gemma-27B       & $\uparrow$ 2.9 & $\uparrow$ 0.8 & $\uparrow$ 4.4\phantom{0} \\

\midrule
\rowcolor{lightlightgray}
\multicolumn{4}{c}{\textbf{\textit{Qwen-2.5 Instruct Models}}} \\

Qwen-0.5B       & $\uparrow$ 5.4 & $\downarrow$ 0.7 & $\downarrow$ 0.6\phantom{0} \\
Qwen-1.5B       & $\uparrow$ 5.8 & $\downarrow$ 0.4 & $\downarrow$ 8.0\phantom{0} \\
Qwen-3B         & $\uparrow$ 4.6 & $\downarrow$ 0.4 & $\uparrow$ 23.7 \\
\textbf{Qwen-7B (self)}  & $\uparrow$ 8.1 & $\uparrow$ 0.8 & $\uparrow$ 21.4 \\
Qwen-14B        & $\uparrow$ 6.7 & $\downarrow$ 0.7 & $\uparrow$ 33.0 \\
Qwen-32B        & $\uparrow$ 2.1 & $\uparrow$ 1.8 & $\downarrow$ 8.3\phantom{0} \\
Qwen-72B        & $\uparrow$ 3.8 & $\downarrow$ 1.2 & $\downarrow$ 24.3 \\
\bottomrule
\end{tabular*}
\caption{Cross-model transfer of FLEx summaries to a fixed target model (Qwen-7B). For each dataset, summaries are distilled from the source model's errors on that dataset and applied to a fixed target model (Qwen-7B) evaluated on the same dataset.}
\label{tab:cross_model}
\end{table}

\end{document}